\documentclass[lettersize,journal]{IEEEtran}
\usepackage{amsmath,amsfonts}
\usepackage{color}
\usepackage{graphicx}
\usepackage{multirow}
\usepackage{algorithmic}
\usepackage{algorithm}
\usepackage{array}
\usepackage{diagbox}
\usepackage{subfigure}
\usepackage{textcomp}
\usepackage{booktabs}
\usepackage{stfloats}
\usepackage{url}
\usepackage{verbatim}
\usepackage{bm}

\usepackage{amssymb}
\usepackage{pifont}
\newcommand{\cmark}{\ding{51}}%
\newcommand{\xmark}{\ding{55}}%

\usepackage{cite}
\begin{document}

\title{Multi-step Temporal Modeling\\ for UAV Tracking}

\author{Xiaoying Yuan, Tingfa Xu$^{\ast}$, Xincong Liu, Ying Wang,  Haolin Qin, Yuqiang Fang and Jianan Li$^{\ast}$\thanks{*Corresponding author}

\IEEEcompsocitemizethanks{\IEEEcompsocthanksitem 
Xiaoying Yuan, Tingfa Xu, Xincong Liu , Ying wang, Haolin Qin and Jianan Li are with Beijing Institute of Technology, Beijing 100081, China (e-mail: \{3120210585, ciom\_xtf1, 3120210514, 3120215325, 3120225333,lijianan \}@bit.edu.cn).
\IEEEcompsocthanksitem
Yuqiang Fang is with the Science and Technology on Complex Electronic System Simulation Laboratory, Space Engineering University, Beijing 101416, China (e-mail: fangyuqiang@nudt.edu.cn).
\IEEEcompsocthanksitem  

Jianan Li and Tingfa Xu are also with the Key Laboratory of Photoelectronic Imaging Technology and System, Ministry of Education of China, Beijing 100081, China 
\IEEEcompsocthanksitem 
Tingfa Xu is also with Chongqing Innovation Center, Beijing Institute of Technology, Chongqing 401135, China.}}




\maketitle

\begin{abstract} 
In the realm of unmanned aerial vehicle (UAV) tracking, Siamese-based approaches have gained traction due to their optimal balance between efficiency and precision. However, UAV scenarios often present challenges such as insufficient sampling resolution, fast motion and small objects with limited feature information. As a result, temporal context in UAV tracking tasks plays a pivotal role in target location, overshadowing the target's precise features. 
In this paper, we introduce MT-Track, a streamlined and efficient multi-step temporal modeling framework designed to harness the temporal context from historical frames for enhanced UAV tracking. This temporal integration occurs in two steps: correlation map generation and correlation map refinement. Specifically, we unveil a unique temporal correlation module that dynamically assesses the interplay between the template and search region features. This module leverages temporal information to refresh the template feature, yielding a more precise correlation map. Subsequently, we propose a mutual transformer module to refine the correlation maps of historical and current frames by modeling the temporal knowledge in the tracking sequence. This method significantly trims computational demands compared to the raw transformer. The compact yet potent nature of our tracking framework ensures commendable tracking outcomes, particularly in extended tracking scenarios.
Comprehensive tests across four renowned UAV benchmarks substantiate the superior efficacy of our approach, delivering real-time performance at 84.7 FPS on a single GPU. Real-world test on the NVIDIA AGX hardware platform achieves a speed exceeding 30 FPS, validating the practicality of our method.

\end{abstract}


\section{Introduction}\label{intro}
\IEEEPARstart{I}{n} recent years, the swift advancements in unmanned aerial vehicles (UAVs) have propelled their widespread use in remote sensing\cite{agri}, intelligent transportation\cite{road,2020Boundary}, geographical survey\cite{9583266} and video surveillance\cite{5229254}. This surge is attributed to their multifunctional capabilities, convenience and efficiency. However, UAV tracking scenarios often grapple with challenges such as small objects, low resolution, occlusion, scale changes and fast motions\cite{fu2021correlation} due to the large field of view and jitter of the UAV's camera. Given the computational constraints of UAVs, ensuring robust, precise, and real-time tracking in intricate environments remains a formidable task for UAV tracking.

\begin{figure}
\centering
\subfigure[Visualization of the original UAV images.]{
\includegraphics[scale=0.21]{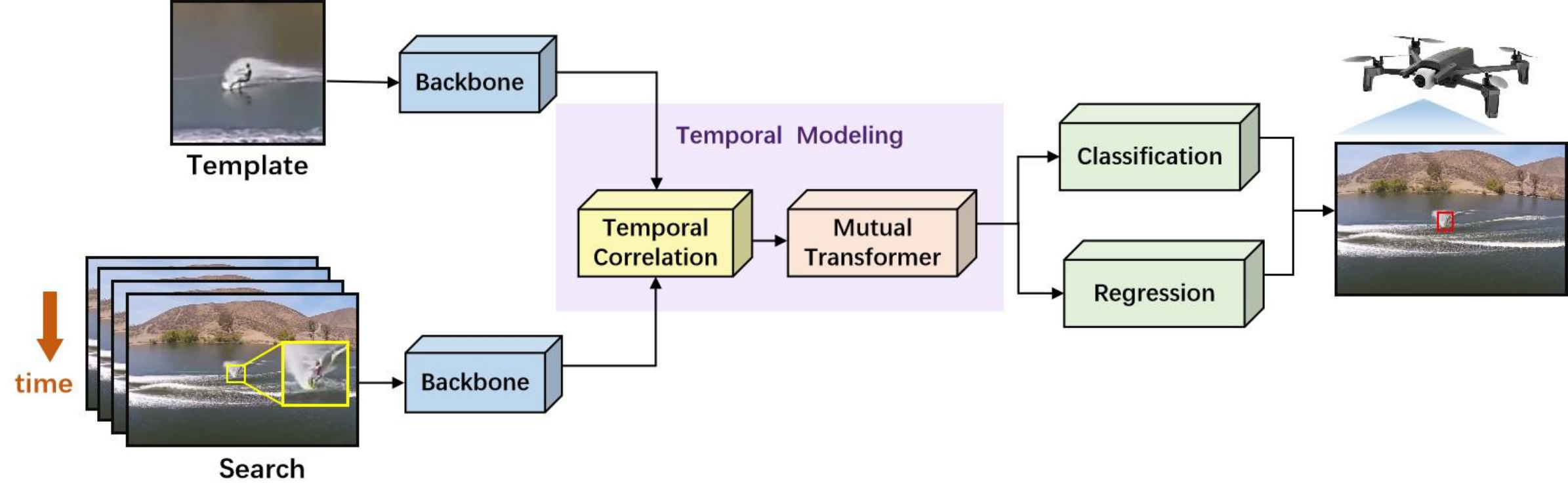}}
\subfigure[Accuracy-speed trade-off on DTB70]{
\hspace{12mm}\includegraphics[width=0.78\linewidth]{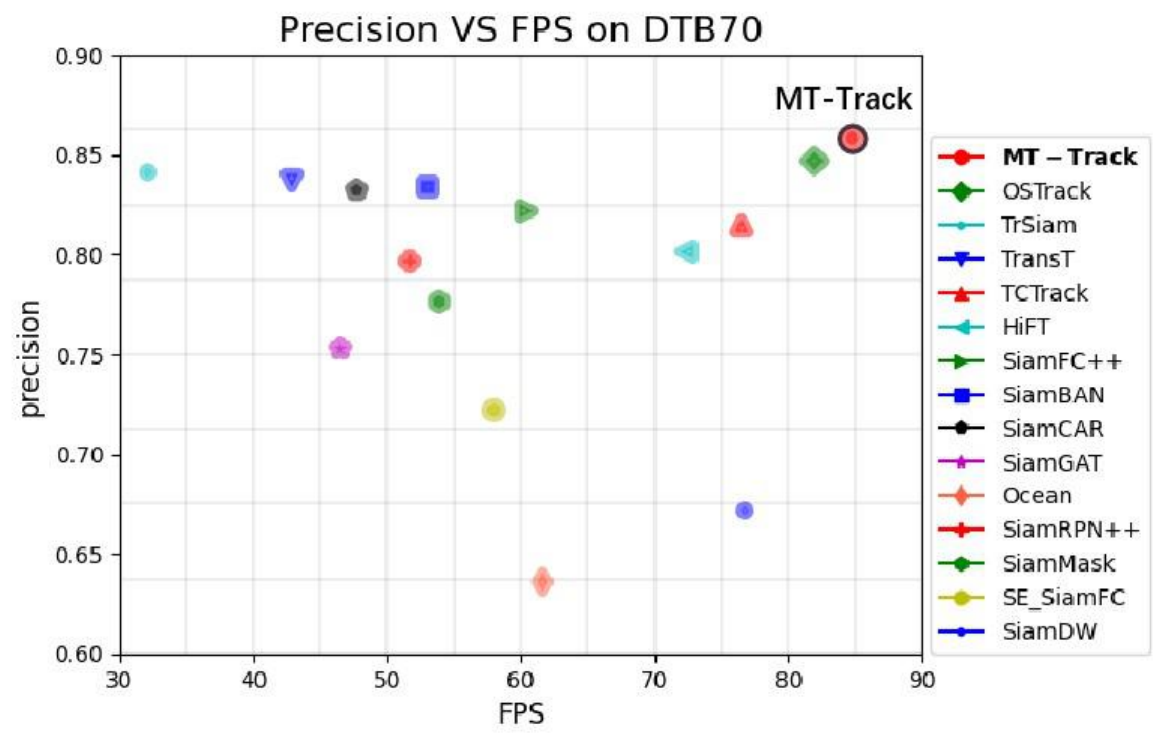}}
\caption{\textbf{(a)  Overview of MT-Track.} The proposed temporal modeling module exploits temporal information at two steps: correlation map generation by temporal correlation, and correlation map refinement by mutual transformer. \textbf{(b) Accuracy-speed trade-off on DTB70.} Our MT-Track achieves competitive performance with impressive efficiency due to the full use of temporal information.}
\label{fig1}
\end{figure}

Recent studies on Siamese-based networks have paved the way for proficient Siamese trackers\cite{SiamAPN++,SiamAPN,Hift}, aligning with the real-time operational demands of UAVs. These trackers employ an initial template to gauge the similarity between the template and search region, utilizing various correlation variants like depth-wise correlation\cite{Ocean, SiamRPN++}, point-wise correlation\cite{9577854}, and naive correlation\cite{SiamFC}. A notable limitation, however, is their underutilization of the abundant temporal data during the tracking phase. This oversight can hinder the tracker's ability to discern target motion shifts, making it susceptible to failures amidst drastic target appearance alterations and occlusions.
\IEEEpubidadjcol
While many prevailing strategies in general object tracking incorporate template updating mechanisms to infuse temporal data via transformers\cite{TMT,HCAT,guo2023vit}, memory networks\cite{1,STMTrack}, concatenation operations\cite{Stark}, and graph networks\cite{graph}, they fall short in UAV tracking contexts. These approaches predominantly involve temporal modeling of pixel-level image features, demanding high image quality and presenting low efficiency, rendering them impractical for deployment on UAV platforms. Moreover, while these methods excel in scenarios with stationary or slow-moving cameras, they falter in UAV-specific challenges like low resolution, fast motion, appearance changes, and small objects as shown in Fig.\ref{fig1} (a). These characteristics of UAV images make the feature information of the target insufficient or damaged during the tracking process, so the methods of focusing on specific feature representations of the target in general tracking is not applicable in UAV tracking. 

To sum up, the key issues in the field of UAV object tracking is how to effectively utilize the limited feature in UAV scenarios and perform efficient temporal modeling to address unique challenges and achieve deployment on hardware platforms.
Addressing these challenges, we introduce MT-Track, a streamlined multi-step temporal modeling framework tailored for UAV tracking as shown in Fig.\ref{fig1} (a). The architecture encompasses three primary modules: feature extraction, multi-step temporal modeling, and target localization. The multi-step temporal modeling extensively exploits the temporal information of video sequences. This modeling approach is based on correlation maps, which not only enhances the efficiency of temporal modeling but also reduces reliance on target pixel-level features. It is particularly suitable for UAV missions.

Concretely, the multi-step temporal modeling module extracts temporal information in two steps as shown in Fig. \ref{fig2} (a).
\textbf{Step 1: Correlation map generation.} Temporal correlation is used to model the relationship between the template and the search region dynamically. First, we propose a Multi-Template Fusion (MTF) module to generate the updated template feature by fusing the historical template feature sequences into the current template feature.
A depth-wise correlation is then utilized to embed the two branches and obtain the correlation map. In this way, we generate more accurate correlation maps by incorporating historical information, distinguishing foreground from background. 
\textbf{Step 2: Correlation map refinement.} Mutual transformer mutually refines the historical correlation map and the current correlation map by modeling the temporal knowledge in the sequence, which contains parallel historical branch and current branch.
Inspired by the raw transformer architecture\cite{Transformer}, our encoder produces the encoded correlation map, which focuses more attention on the target. 
For decoder, we propose a mutual attention (MA) to refine the correlation maps of the current and historical branches by interacting the information from the both sides, which achieves the construction of temporal connections between consecutive frames.

We extensively evaluate our MT-Track on four popular aerial tracking benchmarks, including DTB70\cite{dtb}, UAV123\cite{UAV123}, UAV123@10fps\cite{UAV123}, and UAVTrack112\_L\cite{9477413}. 
Our MT-Track outperforms the top-performing TCTrack\cite{TCTrack} by $4.4\%$ in precision on DTB70 and runs at an impressive real-time speed of 84.7 FPS (Fig. \ref{fig1} (b)), achieving an ideal trade-off between accuracy and speed. The real-world deployment on NVIDIA Jetson AGX Xavier demonstrates the remarkable stability and robustness of MT-Track in UAV object tracking, running at a speed of over 30 FPS.
The contributions of this work can be summarised as follows:

\begin{itemize}
\item We propose a concise but effective multi-step temporal modeling module dedicated to UAV tracking. It is capable of utilizing rich temporal information of video sequences and achieves real-time speed.

\item The unveiling of a novel temporal correlation, enhancing correlation map accuracy through template feature updates.

\item Based on the temporal correlation, the mutual transformer refines the correlation maps by modeling the temporal relationship between consecutive frames using finite features in the UAV scenarios.
Compared to the raw transformer, this mechanism significantly trims computational demands while adeptly harnessing temporal contexts.


\end{itemize}

\section{Related Work}\label{related work}
\subsection{UAV Object Tracking Methods}

The complex environments and hardware limitations associated with unmanned aerial vehicle (UAV) object tracking tasks have placed higher demands on UAV object tracking methods. Currently, mainstream UAV object tracking methods can be primarily classified into two categories\cite{han2022comprehensive}: Discriminative Correlation Filters (DCF)-based trackers \cite{9009064,kiani2017learning,9447987,6,8578162,Han} and Deep Learning-based methods.\cite{zhu2017uct,yuan2023active,SiamAPN,SiamRPN,DaSiamRPN,SiamRPN++,yuan2023robust}.
TB-BiCF\cite{6} proposes a bidirectional incongruity-aware correlation filter that utilises a temporary block to store temporary inter-frame information. Chenwei Deng {et.al} \cite{9447987} introduce dynamic spatial regularization weights in DCF tracking models to encourage filters to focus on more reliable regions during the training phase.
DCF-based trackers can achieve efficient competitive performance due to fast computation in the Fourier domain. However, these trackers struggle to maintain robustness under a variety of tracking conditions due to poor representation of hand-crafted features.

In recent years, Siamese-based trackers \cite{9064815,SiamAPN,SiamRPN,SiamRPN++,DaSiamRPN} have become attractive due to their excellent performance and efficiency.
SiamRPN \cite{SiamRPN} combines a Siamese subnetwork in tracking and region proposal subnetwork network in detection to achieve end-to-end training, achieving a speed of 160 FPS. FASNT\cite{9064815} addresses the problem of large-scale pose variations of UAV objects by proposing a feature alignment module to calibrate the search image. The Siamese-based trackers lay a solid foundation for subsequent UAV tracking, since they provide a good trade-off between efficiency and accuracy. 
However, as these methods do not make full use of temporal information, most of them have difficulty in robustly classifying objects, especially in the presence of interference targets and appearance changes.
\subsection{Transformer in Tracking}

Transformer is first proposed in \cite{Transformer} as a paradigm for machine translation, which has the ability to model the global relationship among the inputs by taking sequences as inputs and learning their dependency. 
Based on its excellent performance, lots of work has applied transformer to a wide range of computer vision tasks, such as image classification \cite{VIT}, object detection \cite{DETR}, and object tracking \cite{Transformertracking,Hift,9914676,10144283,blatter2023efficient,guo2023vit}.

TransT\cite{Transformertracking} proposes an ego-context augment module and a cross-feature augment module to achieve feature fusion, which are based on self-attention and cross-attention. OSTrack\cite{ostrack} proposes a one-stream tracking framework based on Vision Transformer\cite{VIT}.
They achieved a high degree of accuracy in general object tracking, while they do not perform well when faced with the challenge of limited and highly variable feature representation of UAV object characteristics, and computation are too large to apply on UAVs. 
Hift\cite{Hift} proposes a hierarchical transformer architecture, which takes correlation maps as inputs and utilizes cross-attention to fuse correlation maps from different feature levels and obtains high-quality feature representations.
However, appearance changes in the video scenarios are not well considered and the relationships across frames are also be ignored.
In this work, we fully consider the temporal information in sequences and achieve real-time tracking speed.

\subsection{Temporal Modeling}
\begin{figure*}[htp]
\centering
\includegraphics[scale=0.27]{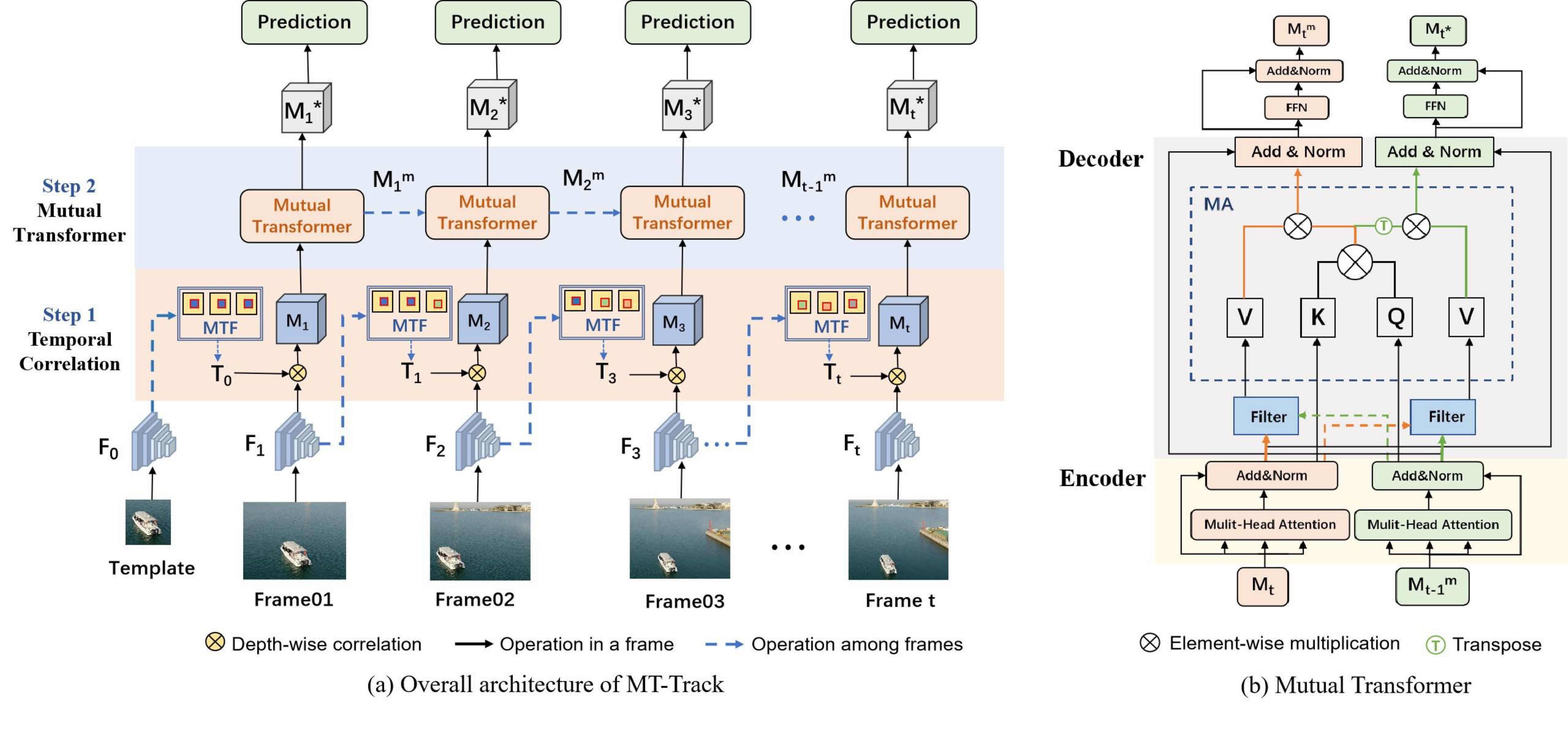}  
\caption{\textbf{(a) Overall architecture of MT-Track.} First, the features extracted from the backbone are fed into temporal correlation to dynamically update template feature and produce accurate correlation maps. The Multi-Template Fusion module (MTF) is described in detail in Fig. \ref{temcor}. Next, mutual transformer accepts the correlation maps to model temporal knowledge, and produces the refined correlation maps. \textbf{(b) Mutual Transformer.} It is an encoder-decoder architecture with parallel historical and current branches. The mutual attention (MA) is introduced to mutually refine the correlation maps.} 
\label{fig2}    
\end{figure*}
Tracking task is actually a sequence-level task, and temporal context is important for tracking tasks, especially when dealing with the challenges of UAV videos. Some methods have been conducted to consider the temporal information \cite{Stark,UpdateNet,TCTrack,9448479,9417705,guo2023vit,yuan2022learning,3,yuan2022aligned,1}. 
STARK\cite{Stark} captures the global dependency in temporal dimension by inputting a triplet consisting of the current search image, the initial template and a dynamic template. TrSiam\cite{TMT} propagates temporal context in template patches by using transformer. UpdateNet\cite{UpdateNet} updates the object template based on the first frame and uses all previous frames to accumulate the template based on the input training tracking data.
These trackers achieve temporal modeling by updating the template features during tracking sequences, while they import temporal information at only one stage, hindering further promotion of tracking performance. ASTCA\cite{yuan2022learning} proposes an adaptive spatial-temporal context-aware model that accurately distinguishes between target and
background in UAV tracking scenarios. TCTrack\cite{TCTrack} proposes a concise framework to explore the contextual information at features level and similarity maps level, which shows competitive effectiveness and efficiency on aerial tracking task.

In this work, we take complete account of temporal information by proposing a multi-step temporal modeling. Except for feature extraction level as TCTrack\cite{TCTrack}, we propose temporal correlation and mutual transformer to achieve temporal modeling in correlation map generation and correlation map reinforcement.

\section{Multi-step Temporal Modeling}

In this section, we describe our Multi-step temporal modeling module in detail, and the overall structure is shown in Fig. \ref{fig2} (a). First, features extracted from the backbone are fed into temporal correlation to dynamically update template feature and generate accurate correlation maps. Next, the correlation maps are fed into mutual transformer to model the temporal knowledge by establishing temporal connections between the frames. In this way, we obtain refined correlation maps that incorporate temporal information from successive frames by implementing temporal modeling in multi-step.
\subsection{Temporal Correlation}
\label{Temporal Correlation}

\begin{figure}[t]
\centering
\includegraphics[scale=0.33]{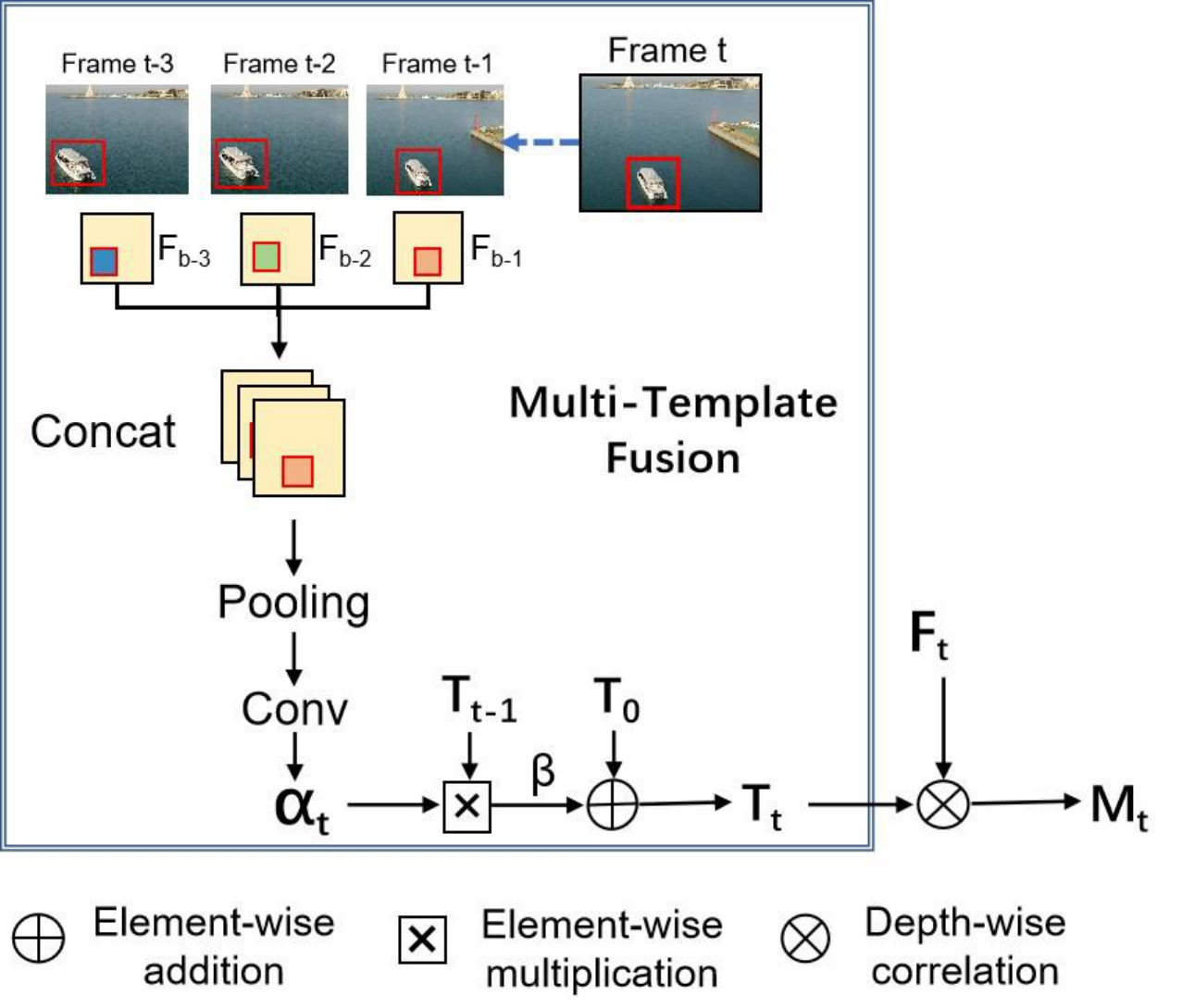}  
\caption
{\textbf{Workflow of Temporal Correlation.} The Multi-Template Fusion (MTF) updates the template feature by fusing the temporal information of the feature sequences. The updated template feature is then used to create the correlation map by depth-wise correlation.}  
\label{temcor}    
\end{figure}

The correlation map in tracking task distinguishes the foreground from the background in the search frame, reflecting the similarity between the template feature and the search region feature, usually calculated by depth-wise correlation \cite{han2021connection}. Similarly, we obtain the correlation map by depth-wise correlation, but the difference is that we design a dynamic linear network to dynamically update the template feature to obtain more accurate correlation maps. As shown in Fig. \ref{temcor}, Multi-Template Fusion (MTF) module dynamically updates the template feature by fusing the current template feature with multi-template features of historical frames. This module perceives the motion information of the target objects in the tracking period and thus generates more accurate correlation map.  

\noindent\textbf{Depth-wise Correlation.}
The calculation of correlation map $\bm{M}_t\in\mathbb{R}^{\mathrm{H}\times \mathrm{W}\times \mathrm{C}}$ is as follows:

\begin{equation}
\bm{M}_t = \bm{T}_t * \bm{F}_t.
\end{equation}
where $*$ represents the depth-wise correlation\cite{han2021connection}, $\bm{F}_t\in\mathbb{R}^{\mathrm{H}_s\times \mathrm{W}_s\times\mathrm{C}}$ is the feature of current search region. 

\noindent\textbf{Multi-Template Fusion.}
The updated template feature $\bm{T}_t\in\mathbb{R}^{\mathrm{H}_T\times \mathrm{W}_T\times \mathrm{C}}$ is obtained by Multi-Template Fusion (MTF) module, which is calculated by:
\begin{equation}
\label{Tt}
\bm{T}_t = \bm{T}_0+\bm{\beta}(\bm{\alpha}_t \times \bm{T}_{t-1}).
\end{equation}
where $\bm{T}_0\in\mathbb{R}^{\mathrm{H}_T\times \mathrm{W}_T\times \mathrm{C}}$ is the initial template initialized by the first frame, $\bm{T}_{t-1}\in\mathbb{R}^{\mathrm{H}_T\times \mathrm{W}_T\times \mathrm{C}}$ is the corresponding template feature obtained from reliable previous frame prediction, $\bm{\beta}$ is a learnable parameter that affects how much temporal information is utilized in the tracking process and is initialized to zero, and $\bm{\alpha}_t$ is a dynamic calibration weight calculated as follows: 

\begin{equation}
\label{a_t}
\bm{\alpha}_t = \bm{W}_1(\mathbf{GAP}(\mathbf{Concat}(\bm{F}_{b-3}, \bm{F}_{b-2}, \bm{F}_{b-1}))).
\end{equation}
where $\mathbf{GAP}$ represents global average pooling and $\mathbf{Concat}$ represents the concatenation.
$\bm{F}_{b-3}$, $\bm{F}_{b-2}$, $\bm{F}_{b-1}$ are the feature maps mapped with the bounding box of the predicted target, generated by $\bm{F}_{t-3}$, $\bm{F}_{t-2}$, $\bm{F}_{t-1}$, which are the feature maps of the three frames closest to the current frame and initialized to the feature map of the first frame $\bm{F}_0$. 
$\bm{\alpha}_t$ is obtained by fusing the target features of the temporal frames, and we employ $\bm{\alpha}_t$ as a dynamic calibration parameter to linearly combine the historical template features with the initial template features to strengthen the parts of the template that are more similar to the changing target and weaken the parts that are not similar. Since the scale and appearance of the target change during the tracking process, $\bm{\alpha}_t$ can effectively correct the initial template.
For the number of frames used in MTF, we also conduct ablation experiments as shown in Section \ref{frame length}.

In MTF module, we control the proportion of influence of $\bm{T}_{t-1}$ through $\bm{\alpha}_t$ and $\bm{\beta}$, because $\bm{T}_{t-1}$ is likely to distort the template during the tracking period, while $\bm{T}_0$ can maintain most of the template information, especially in the face of deformation challenge. 
In this way, the information of initial template is reserved and also adds the temporal contexts, effectively avoiding the template information loss and saving additional selection strategies. 

Additionally, to reduce the impact of prediction failures, we introduce a threshold $\tau$ to select the appropriate template feature. The current feature map is stored in MTF when the confidence score exceeds the threshold. At the same time, the oldest feature map in the memory cache will be popped out.
Our temporal correlation is memory-efficient for that we update the template feature map in the tracking period rather than saving all of them.

\subsection{Mutual Transformer}

For the generated correlation map, we further refine it by establishing the temporal contextual relationship between consecutive frames, as depicted in Step 2 of Fig.\ref{fig2} (a). Specifically, we propose a novel mutual transformer consisting of an encoder and a decoder divided into a historical branch and a current branch as shown in Fig. \ref{fig2} (b).
The encoder leverages self-attention block to learn the relationship between the inputs through their feature dependencies, which makes the tracker pay more attention to the target. 
In the decoder, we propose an innovative Mutual Attention (MA) mechanism where the two branches set different query, key and value for the interaction between the historical correlation map and the current correlation map to obtain refined correlation maps.

\noindent\textbf{Transformer Encoder.}
The transformer encoder consists of Multi-Head Attention ($\mathbf{MHA}$), following\cite{Transformer}. Starting from the basic attention module in $\mathbf{MHA}$, we have:
 \begin{equation}\label{attention}
\mathbf{Att}(\bm{Q}, \bm{K}, \bm{V}) = \mathbf{Softmax}(\frac{\bm{Q}\cdot \bm{K}^T}{\sqrt{d_k}})\cdot \bm{V}.
\end{equation}
where $\bm{Q}\in\mathbb{R}^{\mathrm{N}_q\times \mathrm{C}}$, $\bm{K}\in\mathbb{R}^{\mathrm{N}_k\times \mathrm{C}}$, $\bm{V}\in\mathbb{R}^{\mathrm{N}_v\times \mathrm{C}}$ are the query vector, key vector and value vector, which are the parameters of the attention function, and $\sqrt{d_k}$ is the scaling factor.
Introducing the linear projection matrix to the attention module, the $\mathbf{MHA}$ can be defined as follows:

\begin{equation}\label{MHA}
\begin{aligned}
\mathbf{MHA}(\bm{Q}, \bm{K}, \bm{V}) &= \mathbf{Concat}(\mathbf{Head}_1, ..., \mathbf{Head}_N)\bm{W},\\
\mathbf{Head}_i= \mathbf{Att}&(\bm{Q}\cdot \bm{W}^{Q}_i, \bm{K}\cdot \bm{W}^{K}_i, \bm{V}\cdot{\bm{W}}^{V}_i).\\
\end{aligned}
\end{equation}
where $\bm{W}\in\mathbb{R}^{\mathrm{C}\times \mathrm{C}}$, $\bm{W}^Q_i\in\mathbb{R}^ {\mathrm{C}\times \mathrm{C}_d}$, $\bm{W}^K_i\in\mathbb{R}^{\mathrm{C} \times \mathrm{C}_d}$, $\bm{W}^V_i\in\mathbb{R}^{\mathrm{C}\times \mathrm{C}_d}$. 
$\mathbf{Head}_i$ donates the attention of each head in $\mathbf{MHA}$. In our mutual transformer, we employ $\mathbf{MHA}$ with 6 heads, i.e., $N = 6$ and $\mathrm{C}_d=\mathrm{C}/6$.

The historical branch and the current branch take the correlation map $\bm{M}\in\mathbb{R}^{\mathrm{H}\times \mathrm{W}\times \mathrm{C}}$ generated by the temporal correlation module as input, then output the encoded correlation map as follows: 
\begin{equation}
\begin{aligned}
\bm{M}_t^0 = \mathbf{Norm}(\bm{M}_t&+ \mathbf{MHA}(\bm{M}_t, \bm{M}_t, \bm{M}_t)),\\
\bm{M}_{m-1} = \mathbf{Norm}(\bm{M}_{t-1}^m+& \mathbf{MHA}(\bm{M}_{t-1}^m, \bm{M}_{t-1}^m, \bm{M}_{t-1}^m).
\end{aligned}
\end{equation}
where $\mathbf{Norm}$ represents the layer normalization, $\bm{M}_t$ is the current correlation map and $\bm{M}_{t-1}^m$ is the historical correlation map.

By encoding the correlation map, we can obtain the attention reinforced correlation map, in favor of the subsequent mutual temporal reinforcement operation. 
It is worth noting that we employ the attention operation on the correlation map, which can greatly conserve computational resources, due to the size of correlation map is smaller than the original feature map.

\noindent\textbf{Transformer Decoder.}
The decoder is a variant of the Multi-Head Attention, which takes the encoded historical correlation map $\bm{M}_{m-1}$ and the encoded current correlation map $\bm{M}_t^{0}$ as inputs, then outputs the refined current correlation map $\bm{M}_t^*$ with temporal knowledge and the updated historical correlation map $\bm{M}_t^m$ with information of the latest frame. In this way, it can achieve mutual reinforcement on current map and historical map simultaneously by temporal modeling.

\textit{Mutual Attention.}
As for the historical branch, it takes $\bm{M}_{m-1}^f$ and $\bm{M}_t^{f}$ as inputs, which are obtained by processing $\bm{M}_{m-1}$ and $\bm{M}_t^{0}$ through filter. We use the historical correlation map $\bm{M}_{m-1}^f$ for generating query and the current correlation map $\bm{M}_t^{f}$ for generating key and value, and the attention weights of $\bm{M}_{m-1}^f$ and $\bm{M}_t^{f}$ are applied to $\bm{M}_t^{f}$ to obtain the updated historical correlation map $\bm{M}_t^m$ that combines the information of the current frame. Specially, the attention map $\mathbf{S}$ can be expressed as follows:

\begin{equation}
\mathbf{S}(\bm{M}_{m-1}^f, \bm{M}_t^{f}) = \mathbf{Softmax}(\frac{\bm{M}_{m-1}^f\cdot (\bm{M}_t^{f})^\mathrm{T}}{\sqrt{d_k}}).
\label{map}
\end{equation}

Based on the attention map, we calculate the Multi-Head Attention of the t-th frame following Eq. \ref{MHA}.
Hence, we obtain the output of the historical branch by:

\begin{equation}
\begin{aligned}
\mathbf{Att}(&\bm{M}_{m-1}^f, \bm{M}_t^{f}, \bm{M}_t^{f}) = \mathbf{S}(\bm{M}_{m-1}^f, \bm{M}_t^{f})\cdot \bm{M}_t^{f}, \\
\bm{M}_t^m&=\mathbf{Norm}(\bm{M}_{m-1}^f + \mathbf{MHA}(\bm{M}_{m-1}^f, \bm{M}_t^{f}, \bm{M}_t^{f}).\\
\end{aligned}
\end{equation}
where $\bm{M}_{m-1}^f$ is used to generate the query vector, and $\bm{M}_t^{f}$ is used to generate the key and value. $\bm{M}_t^m$ is the updated historical correlation map that incorporates the information from the latest frame.

For the current branch, the current correlation map $\bm{M}_t^{f}$ is used to generate query, the historical correlation map $\bm{M}_{m-1}^f$ is used to generate key and value, and the attention weights of $\bm{M}_t^{f}$ and $\bm{M}_{m-1}^f$ are applied to $\bm{M}_{m-1}^f$ to obtain the refined current correlation map that combines the information of historical frames. It is worth mentioning that we utilize the transpose operation of the matrix to perform only one matrix multiplication when computing the attention weights between the query and the key, greatly avoiding matrix multiplication. The attention map of the current branch is obtained by:
\begin{equation}
\mathbf{S}(\bm{M}_t^{f}, \bm{M}_{m-1}^f)=\mathbf{S}^\mathrm{T}(\bm{M}_{m-1}^f, \bm{M}_t^{f}).
\end{equation}
Then we obtain the output of the current branch by:

\begin{equation}
\begin{aligned}
\mathbf{Att}&(\bm{M}_t^{f}, \bm{M}_{m-1}^f, \bm{M}_{m-1}^f) = \mathbf{S}^\mathrm{T}(\bm{M}_{m-1}^f, \bm{M}_t^{f})\cdot \bm{M}_{m-1}^f, \\
\bm{M}_t^* &= \mathbf{Norm}(\bm{M}_t^{f} + \mathbf{MHA}(\bm{M}_t^{f}, \bm{M}_{m-1}^f, \bm{M}_{m-1}^f).\\
\end{aligned}
\end{equation}
where $\bm{M}_t^{f}$ is used to generate the query, and $\bm{M}_{m-1}^f$ is used to generate the key and value. $\bm{M}_t^*$ is the refined current correlation map that incorporates the temporal contexts.

\textit{Filter.}
It is worth noting that before the mutual attention, we apply a filter to eliminate the unwanted information to enhance robustness, since there are some interference during tracking period, such as motion blur and occlusion. The calculation of filter is as follows:
\begin{equation}
\begin{aligned}
\bm{M}_t^{f} = &\bm{M}_t^{0} \ast\bm{d}_1(\bm{\omega}),\\
\bm{M}_{m-1}^f = &\bm{M}_{m-1} \ast\bm{d}_2(\bm{\omega}).
\end{aligned}
\end{equation}
where $\bm{d}$ refers to the Chunk function, and  $\bm{\omega}$ is a learnable parameter to extract useful information, which is designed as:
\begin{equation}\label{w}
\bm{\omega} = \mathbf{\sigma}(\bm{W}_2\mathbf{\delta}(\bm{W}_1(\mathbf{Concat}(\bm{M}_t^{f}, \bm{M}_{m-1}^f))).
\end{equation}
where $\mathbf{\sigma}$ refers to Sigmoid function, $\mathbf{\delta}$ refers to $\mathbf{ReLU}$ activation function, $\bm{W}_1\in\mathbb{R}^{2\mathrm{C}\times \frac{\mathrm{C}}{r}}$, $\bm{W}_2\in\mathbb{R}^{\frac{\mathrm{C}}{r}\times 2\mathrm{C}}$ (r is the dimensionality reduction ratio\cite{SE-Net}).

As a result, we incorporate useful current similarity map into the temporal contexts for next target location prediction. In the inference process, each frame is added to the updated temporal knowledge instead of storing them all, which is memory efficient. While in the training process, we choose $L$ frames randomly as the previous frames in every video sequence.




\begin{figure*}
\centering  
\vspace{-0.35cm} 
\subfigtopskip=2pt 
\subfigbottomskip=2pt 
\subfigcapskip=-5pt 
\subfigure{
\includegraphics[width=0.30\linewidth]{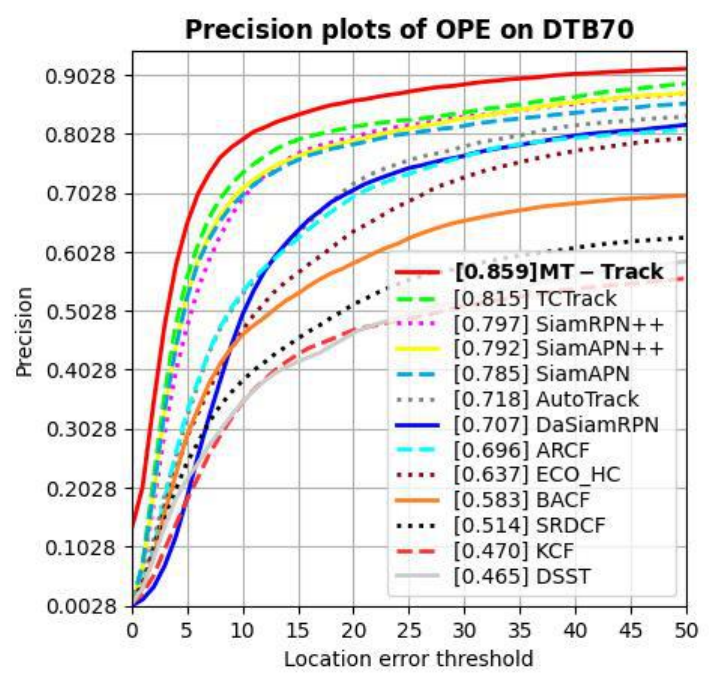}}\hspace{-1mm}
\subfigure{
\includegraphics[width=0.30\linewidth]{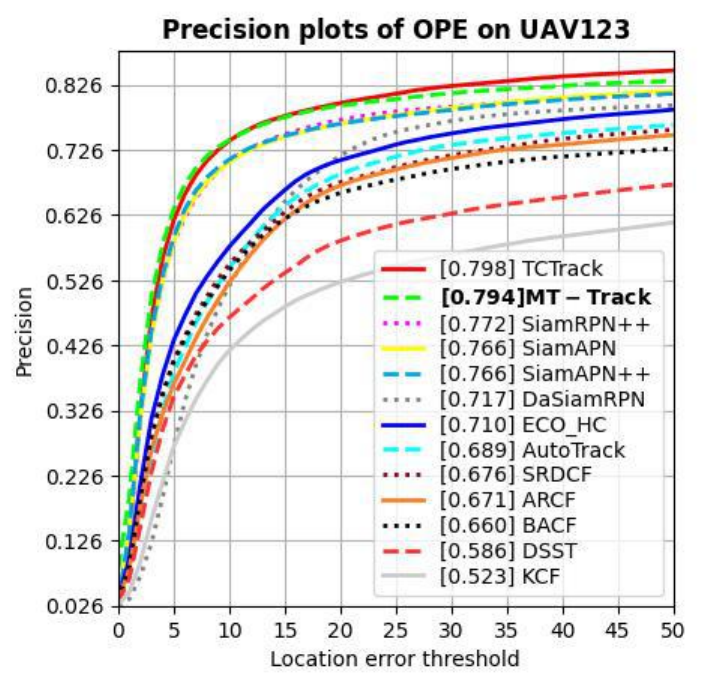}}\hspace{-1mm}
\subfigure{
\includegraphics[width=0.30\linewidth]{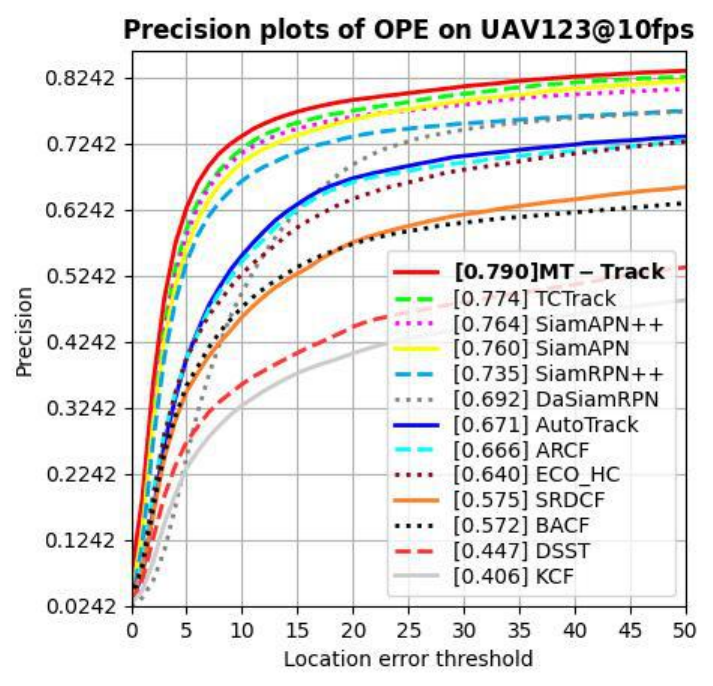}}\\
\subfigure{
\includegraphics[width=0.30\linewidth]{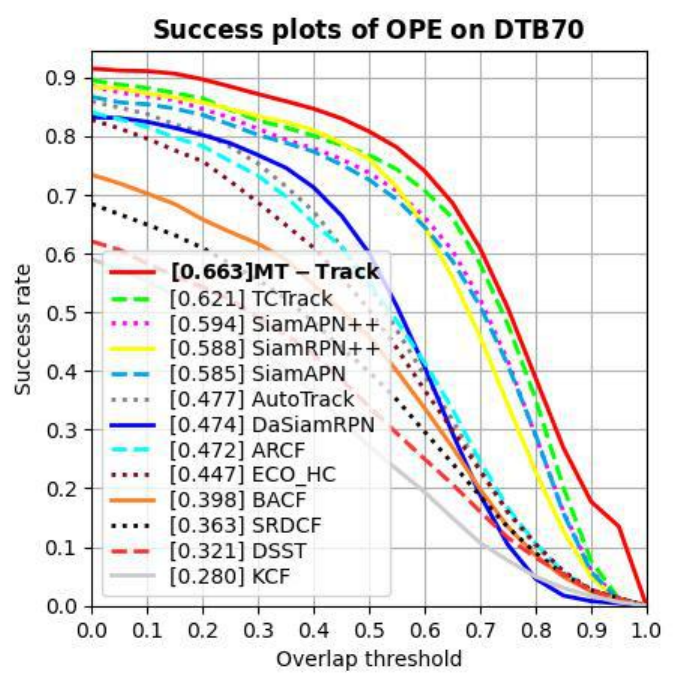}}\hspace{-1mm}
\subfigure{
\includegraphics[width=0.30\linewidth]{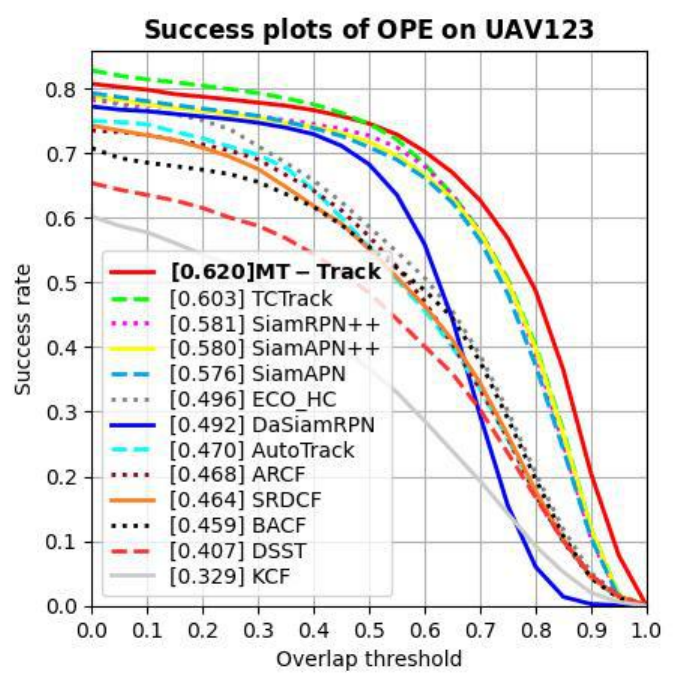}}\hspace{-1mm}
\subfigure{
\includegraphics[width=0.30\linewidth]{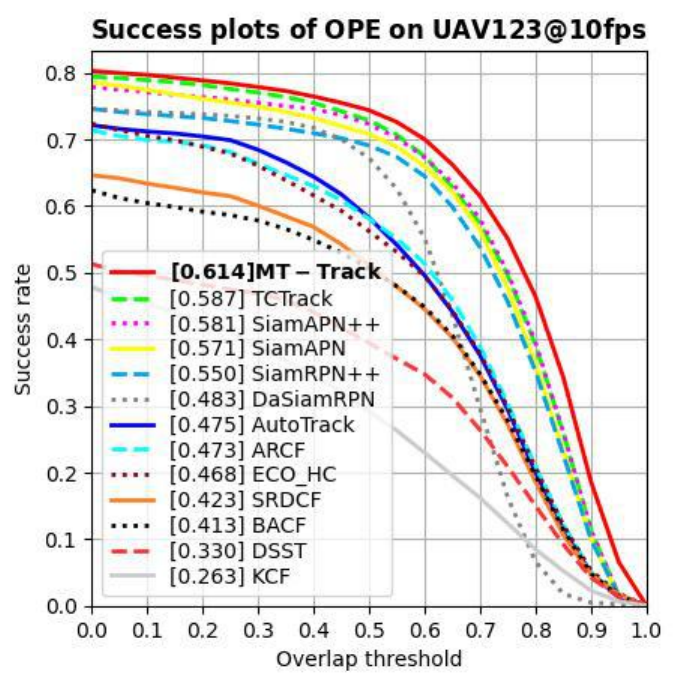}}
\caption{Overall tracking performance. Success and Precision plots of all trackers on DTB70\cite{dtb}, UAV123\cite{UAV123} and UAV123@10fps\cite{UAV123}. Our tracker achieves leading performance against most SoTA trackers.}
\label{vs}
\end{figure*}

\section{Experiments}
In this section, we comprehensively evaluate our proposed MT-Track on four widely-used aerial tracking benchmarks, i.e., DTB70\cite{dtb}, UAV123\cite{UAV123}, UAV123@10fps\cite{UAV123}, and UAVTrack112\_L\cite{9477413}. Ablation studies are further performed to validate our design and parameter settings.

\subsection{Implementation Details}

We use the online TAdaConv\cite{tada} which uses AlexNet with ImageNet pre-trained weights as backbone to extract features. 
The online TAdaConv integrates temporal contextual information in feature extraction and has low latency and maintains high inference efficiency. 
In this work, we initialized the online TAdaConv as in \cite{TCTrack}.
The mutual transformer has one encoder layer and two decoder layers. As for the dimensionality reduction ratio in the filter, we choose 2 ($r=2$). 

\noindent\textbf{Training.}
We train our model on four training sets, including LaSOT\cite{lasot}, GOT-10K\cite{got} and VID\cite{VID}. The size of search images and templates are $287\times287$ pixels and $127\times127$ pixels, respectively. The whole training process of MT-Track consists of 100 epochs. 
For the first 10 epochs, we train our network expect for backbone, and the parameters of layers in backbone are frozen as\cite{SiamRPN++}. The rest of the training processes employ a learning rate that decays from $5e-3$ to $5e-4$ in log space. We employ the stochastic gradient descent (SGD) as the optimizer with a momentum of 0.9, where the mini-batch is 100 pairs.
In the training process, we select two frames with a maximum frame index difference of 100 from each video as the target template and the search region. Furthermore, the length of historical frames is set to 3 ($L=3$).

\noindent\textbf{Inference.}
For inference, we set the threshold $\tau=3.0$ to filter out unhelpful historical frames according to the inference confidence. The historical frames are considered reliable if the score is higher than the threshold $\tau$. In the mutual transformer, we set the initial historical correlation map to the correlation map of the first frame.
All the testing process is implemented on NVIDIA GeForce RTX 3090 Ti. The tracking speed of 84.7 FPS proves the efficiency of MT-Track.

\subsection{Quantitative Results}

In this section, we comprehensively compare MT-Track with existing 12 efficient trackers on four challenging UAV tracking benchmarks, including TCTrack\cite{TCTrack}, HiFT\cite{Hift}, SiamRPN++ \cite{SiamRPN++}, DaSiamRPN\cite{DaSiamRPN}, SiamAPN++ \cite{SiamAPN++}, SiamAPN \cite{SiamAPN++}, ECO \cite{danelljan2017eco}, AutoTrack \cite{li2020autotrack}, DSST \cite{7569092}, BACF \cite{kiani2017learning}, SRDCF \cite{7410847}, KCF \cite{6870486} and ARCF \cite{9009064}.
For fairness, all the Siamese-based trackers adopt the same backbone, i.e., AlexNet pretrained on ImageNet \cite{russakovsky2015imagenet}.

\noindent\textbf{Results on DTB70}.
DTB70\cite{dtb} includes 70 sequences concentrating on severe camera motion in various challenging scenes. 
We adopt DTB70 to verify the robustness of the proposed method and evaluate the effectiveness of our MT-Track in dealing with motion challenge.
As shown in Fig. \ref{vs}, the results clearly show that our method surpasses previous methods to some extent. 
Concretely, our MT-Track achieves 0.859 in precision and 0.663 in area-under-the-curve (AUC), which ranks 1st with an improvement of $4.2\%$ (precision) and $4.4\%$ (AUC) against the other best trackers. This evidences that MT-Track perceives appearance change and state change of the target more accurately, where temporal correlation plays an important role.

\noindent\textbf{Results on UAV123}.
UAV123\cite{UAV123} contains 123 challenging video sequences with more than 112K frames, which are captured from unmanned aerial vehicles at low altitude and have the nature of low-resolution motion blur.
The tracking performance evaluated on UAV123 can prove the capacity of accurate locating in most commonly UAV tracking situations. As shown in Fig. \ref{vs}, our MT-Track achieves 0.620 in AUC and 0.794 in precision, which surpasses all other SoTA trackers in terms of AUC. The performance proves that our multi-step temporal modeling module is effective in UAV tracking.

\begin{table}[t]
\renewcommand{\arraystretch}{1.2}
\setlength{\tabcolsep}{20.0pt}
\caption{Performance comparison on UAVTrack112\_L. The best result is shown in \textbf{bold}. \label{uav112l}}
\centering
\begin{tabular}{l| c c }
\toprule
Method  & Pre.  & Suc.\\
\midrule
ARCF\cite{9009064} & 0.640 & 0.399\\
BACF\cite{kiani2017learning} & 0.593 & 0.358 \\
SRDCF\cite{7410847} & 0.508 & 0.320 \\
DSST\cite{7569092} & 0.619 & 0.386 \\ 
ECO \cite{danelljan2017eco} & 0.684 & 0.436 \\
AutoTrack\cite{li2020autotrack} & 0.675 & 0.405 \\ 
DaSiamRPN\cite{DaSiamRPN} & 0.729 & 0.479 \\
SiamRPN++\cite{SiamRPN++} & 0.773 & 0.479 \\
Hift\cite{Hift} & 0.734 & 0.551 \\
TCTrack\cite{TCTrack} & 0.786 & 0.582 \\ 
\midrule
\textbf{MT-Track (ours)} & \textbf{0.801} & \textbf{0.617} \\
\bottomrule
\end{tabular}
\end{table}

\noindent\textbf{Results on UAV123@10fps}.
Consisting of 123 sequences, UAV123@10fps\cite{UAV123} is down-sampled from a 10 FPS version of sequences.
As the frame interval increases, the challenge of many scenarios becomes more challenging, such as fast camera motion and low resolution. 
Accordingly, to fully asses the capability of trackers in the face of fast motion, UAV123@10fps is more appropriate than UAV123. 
From the comparison with other SoTA trackers, our MT-Track achieves 0.614 in AUC and 0.790 in precision. 
We can obviously observe that our MT-Track possesses superior performance and outperforms the second-best tracker in terms of AUC ($2.7\%$) and precision ($1.6\%$).

\newcommand{\tabincell}[2]{\begin{tabular}{@{}#1@{}}#2\end{tabular}} 
\begin{figure*}[t]
\centering
\includegraphics[scale=0.35]{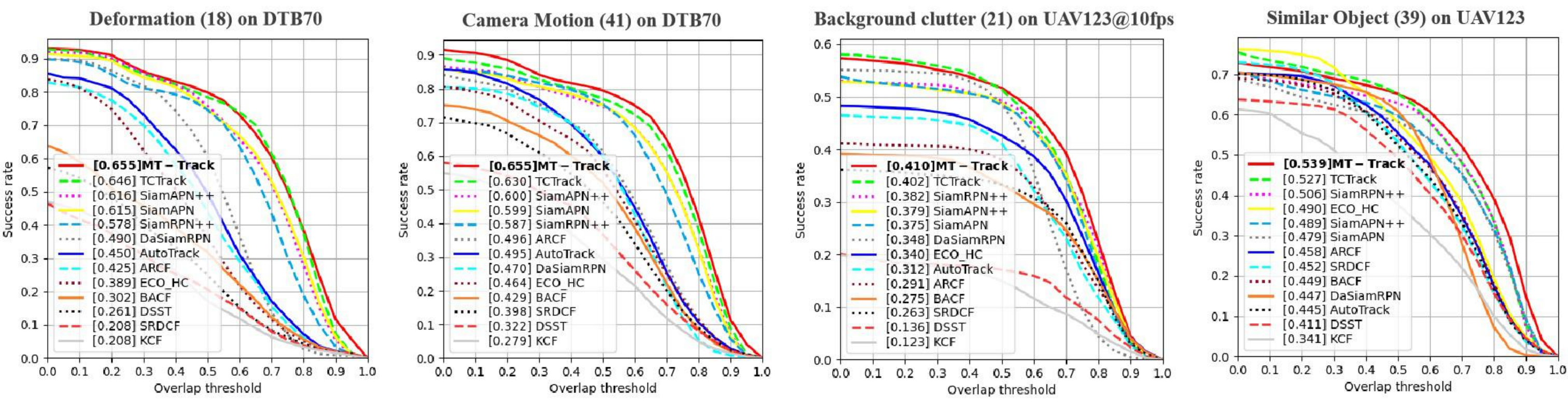}  
\caption
{Attribute-based evaluation. Success plots with attributes of all trackers on DTB70, UAV123, and UAV123\_10fps. It shows MT-Track can maintain promising performance under deformation, camera motion, background clutter, and similar object compared with other 12 trackers.}  
\label{att}    
\end{figure*}

\noindent\textbf{Results on UAVTrack112\_L}.
UAVTrack112\_L\cite{9477413} is the current largest long-term aerial tracking benchmark, containing 112 sequences with various challenging scenes including over 60k frames.
In order to prove the effectiveness of our method in long-term tracking performance, we perform the performance comparison with other trackers on UAVTrack112\_L as shown in Table \ref{uav112l}. 
Thanks to our effective framework that fully models temporal information, our MT-Track demonstrates the best performance against SoTA trackers in terms of precision (0.801) and AUC (0.617), with an improvement of $1.5\%$ on precision and $3.5\%$ on AUC against the second-best tracker. The results fully prove that MT-Track can utilize the temporal information in the long-term video sequence to achieve excellent tracking performance.

\noindent\textbf{Average evaluation on four benchmarks.}
All the four benchmarks mentioned above can reflect the performance of trackers. In order to have a holistic and intuitive comparison, we present the average precision and success rate of the top 10 lightweight trackers on the benchmarks mentioned above. As shown in Table \ref{average}, our MT-Track yields the best precision (0.812) and average success rate score (0.629), which improves the second-best tracker TCTrack\cite{TCTrack} by $1.8\%$ and $3\%$ in precision and success rate, respectively. 
Our model demonstrates a significant improvement over previous work.

\begin{table}[t]
\renewcommand{\arraystretch}{1.2}
\setlength{\tabcolsep}{20.0pt}
\caption{Average evaluation on four aerial tracking benchmarks. Our tracker outperforms all other trackers with an obvious improvement. The best results are shown in \textbf{bold}.\label{average}}
\centering
\begin{tabular}{l| c c }
\toprule
Method  & Pre.  & Suc.\\
\midrule
ARCF\cite{9009064} & 0.669 & 0.453\\
BACF\cite{kiani2017learning} & 0.602 & 0.407 \\
SRDCF\cite{7410847} & 0.568 & 0.393 \\
DSST\cite{7569092} & 0.529 & 0.361 \\ 
ECO \cite{danelljan2017eco} & 0.657 & 0.462 \\
SiamAPN\cite{SiamAPN} & 0.780 & 0.586 \\ 
DaSiamRPN\cite{DaSiamRPN} & 0.711 & 0.482 \\
SiamRPN++\cite{SiamRPN++} & 0.769 & 0.570 \\
SiamAPN++\cite{SiamAPN++} & 0.758 & 0.573\\
Hift\cite{Hift} & 0.779 & 0.580 \\
TCTrack\cite{TCTrack} & 0.794 & 0.599 \\ 
\midrule
\textbf{MT-Track (ours)} & \textbf{0.812} & \textbf{0.629} \\
\bottomrule
\end{tabular}
\end{table}

\subsection{Efficiency.} 
To demonstrate that our tracker can achieve a promising balance between accuracy and efficiency. We compare FLOPs and Params with SoTA trackers as demonstrated in Table \ref{tab:para}. The experiments are executed on NVIDIA GeForce RTX 3090 Ti. FLOPs reflects the time complexity of model, while Params represents the space complexity of model, and both of them can measure the complexity of the model. 

For a more comprehensive comparison, we divide all trackers into two categories: with deeper backbones (on the top) and lightweight backbones (on the bottom). Our MT-Track outperforms all trackers in terms of FLOPS (4.50G) and Parameters (6.10M). For speed analysis, our MT-Track achieves a real-time speed of 84.7 FPS, outperforming all trackers with deeper backbones.
For accuracy, our method achieves the best success rate (0.663) and precision (0.859) compared with all trackers. 
The comparison demonstrates that our method has an excellent balance between accuracy and efficiency, which proves that our method can exploit temporal information in an efficient way and it has the potential to be applied to embedded platforms. 
\begin{table}[]
\renewcommand\arraystretch{1.3}
\centering
\caption{Comparison of FLOPs, parameters, inference speed (FPS), and performance on DTB70 with SoTA trackers. We divide all trackers into two categories: Trackers with deep backbones (Deep) and trackers with lightweight backbones (Light). }
\setlength{\tabcolsep}{1.5mm}{
\begin{tabular}{c|l|ccc|cc}
\toprule
& Method &  \makebox[0.04\textwidth][c]{FLOPs} &  \makebox[0.04\textwidth][c]{\#Params} & \makebox[0.04\textwidth][c]{Speed}  &\makebox[0.04\textwidth][c]{Suc.} &\makebox[0.04\textwidth][c]{Pre.}\\ \midrule
\multirow{7}{*}{Deep} &  OSTrack-256\cite{ostrack} & 21.5 G     & -     & 81.9  & 0.651 & 0.847   \\
& TransT\cite{Transformertracking} & -     & 23.0M     & 42.8  & 0.645 & 0.838   \\
&  SiamRPN++\cite{SiamRPN++} & 48.9 G     & 54.0 M     & 51.6  & 0.615&0.800    \\
&SiamMask\cite{Wang2018SiamMask}  & 16.7 G     & 18.8 M      & 53.8   &  0.575& 0.777 \\
&Ocean\cite{Ocean}      &4.4 G     & 25.9 M  & 61.6  &0.455 & 0.636     \\
&SiamBAN\cite{siamban}   & 48.9 G   & 53.9 M   & 53.0 &  0.643  & 0.834 \\
&SiamCAR\cite{SiameseCAR}   & 48.7 G    & 51.4 M      & 47.4  &0.603 &0.833     \\ \midrule
\multirow{5}{*}{Light}&  DaSiamRPN\cite{DaSiamRPN}    & 21.1 G    & 19.6 M      & \textbf{163.2}  &0.474 &0.707    \\
&SiamAPN\cite{SiamAPN}              & 10.3 G    & 15.1 M      & 127.5  &0.585&0.785    \\
&SiamAPN++\cite{SiamAPN++}            & 9.2 G     & 12.2 M      & 93.4 &0.594&0.792      \\
&TCTrack\cite{TCTrack}  & 4.51 G     & 6.26 M     & 76.4  &0.620 &0.815     \\
\cline{2-7}
&\textbf{MT-Track (ours)}    & \textbf{4.50 G}   &\textbf{6.10 M}     & 84.7 &   \textbf{0.663} &  \textbf{0.859}   \\ \bottomrule
\specialrule{0em}{1pt}{1pt}
\end{tabular}}
\label{tab:para}
\end{table}

\subsection{Attribute-based performance.}

To evaluate our tracker under various challenges, attribute-based comparisons are implemented. Fig. \ref{att} illustrates the comparison of success rate for attribute-based evaluation. Note that MT-Track performs favorably against other 12 trackers in attributes of deformation, similar object, background clutter, and camera motion.

\begin{figure}[t]
\centering
\includegraphics[scale=0.41]{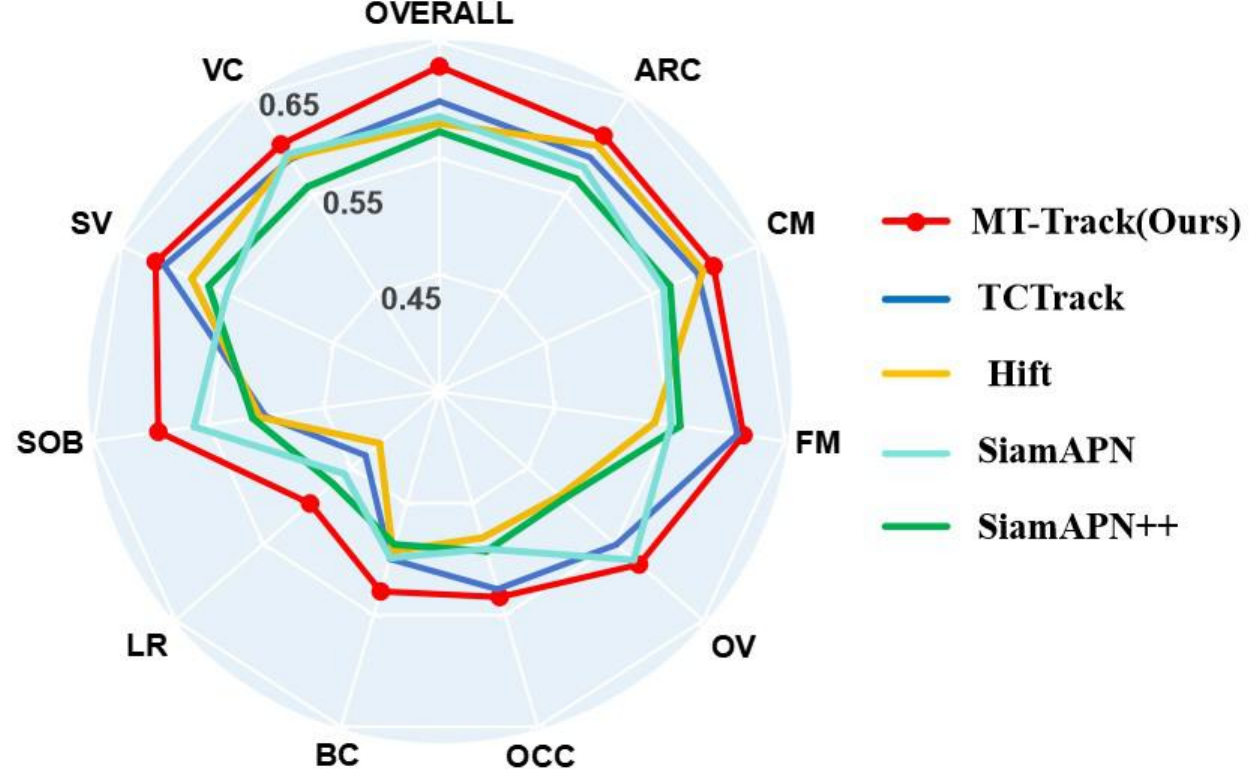}  
\caption
{Average attribute-based evaluation of five real-time Siamese trackers on four UAV tracking benchmarks. This figure compares the average success rate of the trackers when dealing with different challenges and the overall performance. Our MT-Track performs best in different attributes.}  
\label{average_attribute}    
\end{figure}

We also conducted a comprehensive attribute-based performance analysis on four datasets, with the average performance for each attribute determined across all datasets, as illustrated in Fig. \ref{average_attribute}. Our tracker consistently outperforms the top 5 trackers in 10 challenging scenarios, demonstrating its superiority in diverse drone environments. Note that our MT-Track performs well in terms of low-resolution and cluttered background attributes, significantly outperforming the second-ranked tracker by $2.6\%$ and $2.9\%$, respectively. This can be attributed to our multi-step temporal modeling approach based on correlation maps, which skillfully tracks low-resolution targets by reducing reliance on image features and extensive utilization of temporal information.

\begin{figure*}[t]
\centering
\includegraphics[scale=0.25]{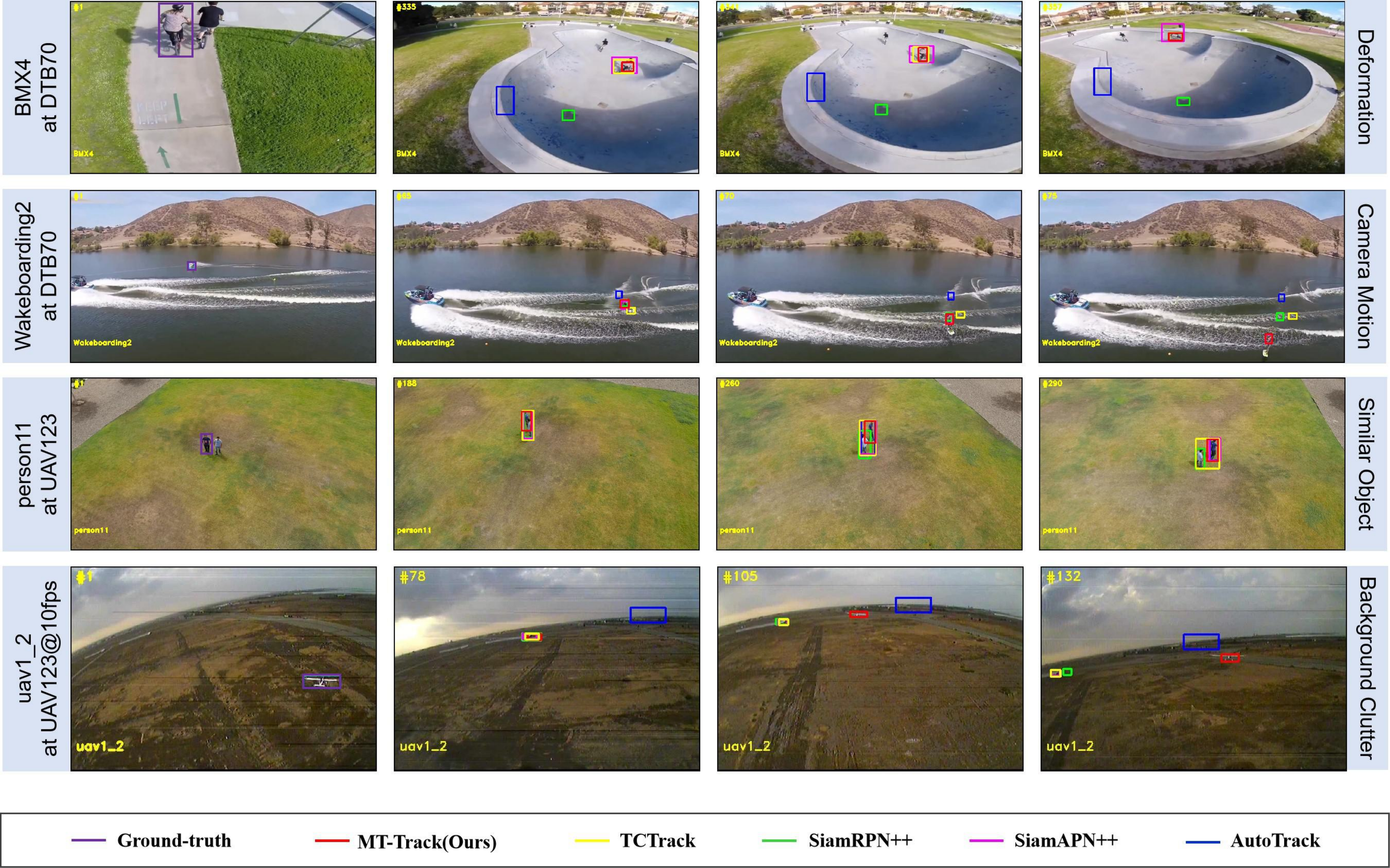}  
\caption
{Qualitative comparison of MT-Track with other four top-performing trackers on some challenging sequences. The sequences and the corresponding benchmarks are BMX4 at DTB70, Wakeboarding2 at DTB70, person11 at UAV123, uav1\_2 at UAV123@10fps. The predicted boxes of different trackers and the ground-truth bounding boxes are indicated in different colors.}  
\label{quality}    
\end{figure*}


\begin{figure*}[tp]
\centering
\includegraphics[scale=0.32]{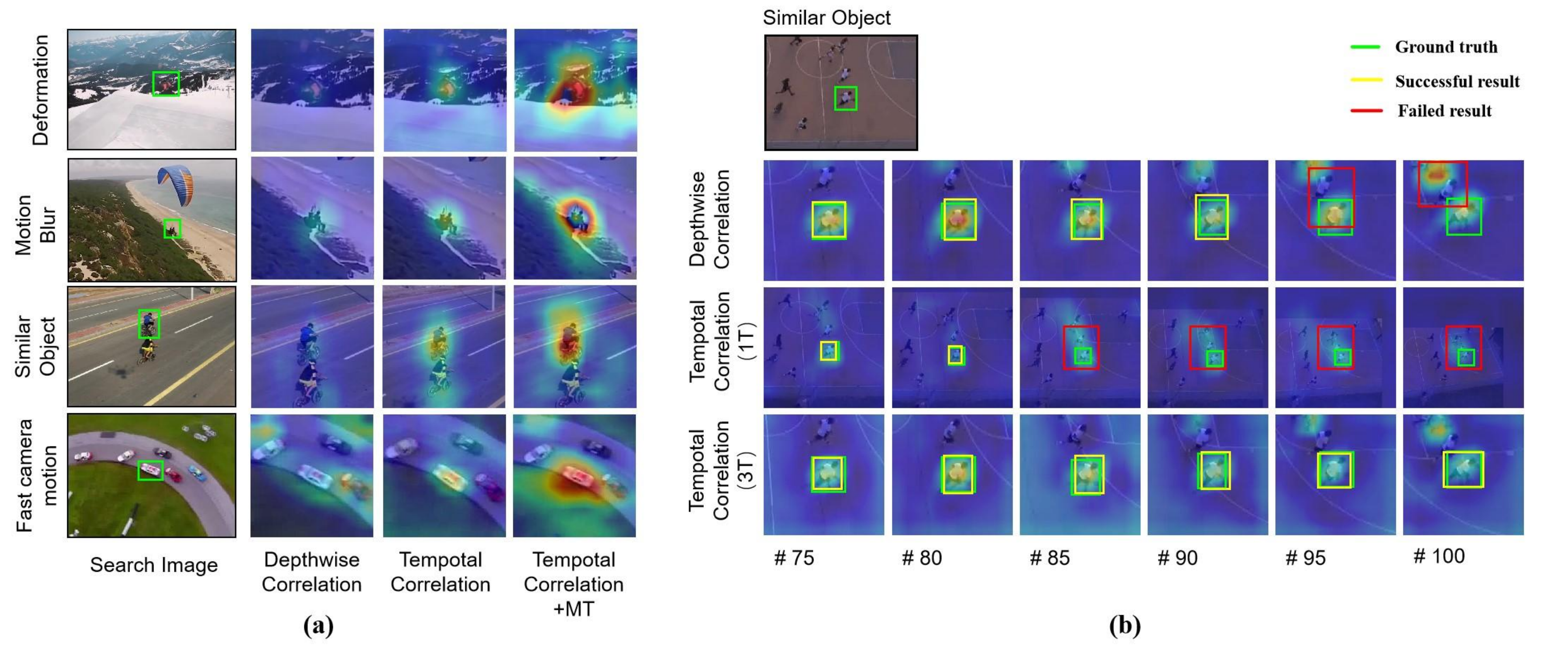}  
\caption
{Visualization of the correlation map.  (a) shows the visualization of the correlation maps produced by different components on some challenging sequences. (b) shows the visualization of different frames when facing similar object, among which the results of the last row are more robust. The objects are marked in red in the search image.}  
\label{vis}    
\end{figure*}

 \subsection{Visualization}
\label{visual}

\noindent\textbf{Qualitative evaluations.} 
In order to show the performance of trackers more intuitively, the qualitative evaluations are shown in Fig. \ref{quality}.
We compare the predicted results of different scenarios from different benchmarks among the top 5 trackers in the quantization results. In each video, we choose three frames with different intervals to present the results, and the ground-truth is shown in the first column. The predicted bounding boxes of MT-Track are most closest to the ground-truth bounding boxes, which proves that MT-Track can always track the right target accurately despite the face of challenging conditions. Benefiting from the proposed multi-step temporal modeling structure, our tracker is more robust in dealing with challenging scenarios such as deformation, camera motion, similar object and background clutter.

\begin{figure}[tp]
\centering
\includegraphics[scale=0.33]{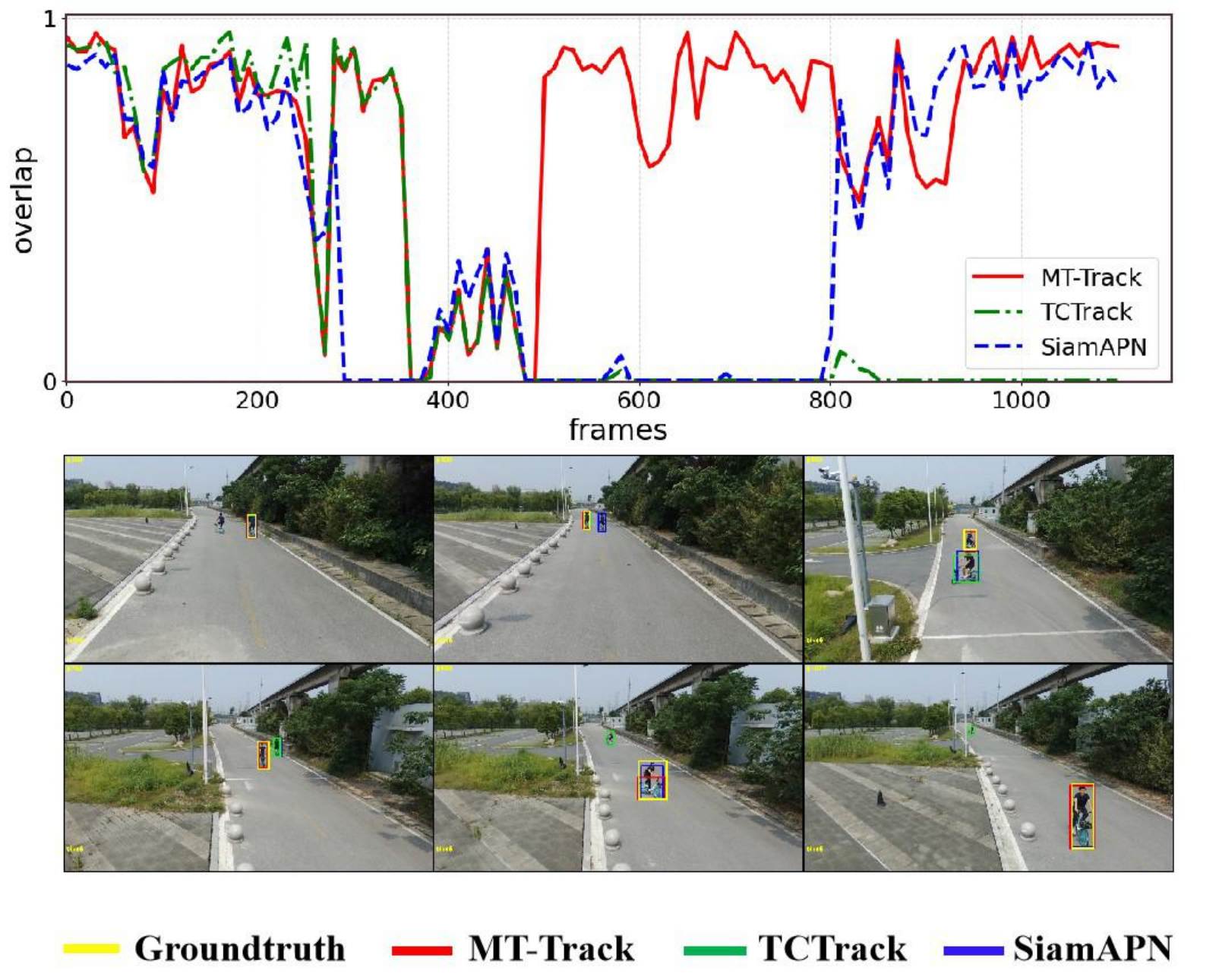}  
\caption
{The tracking results of video bike6 in UAVTrack112\_L. Upper row: overlaps of MT-Track, TCTrack\cite{TCTrack} and SiamAPN\cite{SiamAPN}. The overlaps are defined as intersection-over-union (IOU) between ground-truth bounding box and predicted bounding box. Lower row: tracking snapshots of the three methods. }  
\label{longterm}    
\end{figure}

\noindent\textbf{Performance in long-term tracking.}
Our method has splendid performance in long-term tracking, owing to the full use of temporal information. 
To demonstrate the performance of our method in long-term tracking task, we show the visualization of each frame in a single video sequence on UAVTrack112\_L\cite{9477413} to compare our tracker with TCTrack\cite{TCTrack} and SiamAPN\cite{SiamAPN}. As shown in Fig. \ref{longterm}, the graph above shows the overlap of three trackers in each frame, which helps us to identify the moment when tracking failure begins and ends. 
The graph below shows the tracking results and ground truth of different frames. In contrast to the overlapping results, when intervening targets appear, other models fail, while MT-Track can still locate the correct target due to the temporal modeling module. In addition, our MT-Track can recover quickly after failures, representing its good robustness.

\noindent\textbf{Visualization of correlation map.}
Our MT-Track takes full advantage of temporal information in multi-step temporal modeling. To show the effectiveness of the module obviously, the visualization in Fig. \ref{vis} shows the comparison of correlation maps. 

\textit{Correlation maps from different components.} 
To demonstrate the role of the two components in multi-step temporal modeling, i.e., temporal correlation and mutual transformer, we conduct the experiments on the visualization of correlation maps generated by different components. The results in Fig. \ref{vis} (a) show the comparison of correlation maps in some challenging scenarios, i.e., fast camera motion, similar object, motion blur and deformation. 
Compared to the correlation maps produced by the model with depth-wise correlation, the model with temporal correlation module focuses more attention on the target, and adding mutual transformer (MT) module can highlight the target more explicitly. The results indicate that the multi-step temporal modeling module obtains more effective correlation maps by adequately modeling the temporal information, and thus locates the target more precisely.

\textit{Visualization of interference targets.}
The proposed temporal correlation module perceives the change of template by modeling temporal information, making the model more robust. To verify the robustness of our model, we implement the experiment on the scenarios of similar object.
The visualization in Fig. \ref{vis} (b) illustrates that our tracker can keep focusing on the right object facing to the interference by similar object, while the tracker with general depth-wise correlation easily concentrates on the similar object. 
Specifically, the model with temporal correlation module focuses obvious attention on the interference target at frame \#100 as shown in the second line, while the tracker without temporal correlation shows obvious attention on interference target at frame \#90 as shown in the first line and the inappropriate attention is more than the attention on right one.
Moreover, the attention on the object is scary when the length of previous frames equals 1, which is hard to locate the right object. And turn to Section \ref{n} for a more detailed description.

\subsection{Ablation Study}
To validate the effectiveness of the proposed modules, comprehensive ablation experiments which are conducted on DTB70 are presented in this subsection. We first prove the effectiveness of the components in our framework. Then we analyze the structure and parameter settings of each component.

\noindent\textbf{Effectiveness of Different Components.}
Although the proposed method shows a significant improvement, we would like to further understand the role of our main components and how their combination works, i.e., temporal correlation and mutual transformer. We perform sufficient ablation experiments and present the results in Table \ref{ablation}. 

We design a baseline model that uses TAdaConv (Backbone) and depth-wise correlation, achieving 0.736 in precision and 0.528 in success rate.
First, we replace depth-wise correlation with temporal correlation (Tem-cor) module (shown in the second line), and the performance improves by $1.3\%$ and $0.9\%$ in success rate and precision separately, which proves the effect of our temporal modeling in correlation map generation step.
Next, the mutual transformer (MT) is added to the model with a massive improvement of $10.3\%$ in precision and $12.4\%$ in success rate (shown in the third line).
The significant enhancement proves that mutual transformer can effectively model temporal knowledge to track the object more concisely.
The whole MT-Track, consisting of both temporal correlation and mutual transformer modules leads to more significant performance, raising the performance by about $12.3\%$ in precision and $13.5\%$ in success rate (shown in the last line). 
The results verify that both of the proposed modules improve the tracking performance by temporal modeling, and the combination of them has more significant effect.

\begin{table}[!t]
\caption{Ablation study of different components in MT-Track. Tem-cor: temporal correlation. MT: mutual transformer.}
\centering
\renewcommand{\arraystretch}{1.1}
\setlength{\tabcolsep}{10.0pt}
\begin{tabular}{ c c c|c c}
\toprule 
 \multirow{2}{*}{Tem-cor} & \multicolumn{2}{c|}{MT} & \multirow{2}{*}{Pre.} & \multirow{2}{*}{Suc.} \\ \cline{2-3}
 & Encoder & Filter & &  \\
\midrule
\xmark & \xmark & \xmark  &  0.736 & 0.528\\
\cmark  & \xmark & \xmark &  0.749 & 0.537 \\
\xmark & \cmark & \cmark &   0.839 & 0.652\\
\cmark &\cmark & \xmark & 0.820 & 0.611\\
\cmark & \xmark & \cmark  & 0.792 & 0.584\\
\cline{1-5}
\cmark &\cmark & \cmark  & \textbf{0.859} & \textbf{0.663}\\
\bottomrule
\end{tabular}
\label{ablation}
\end{table}

\begin{figure}[t]
\centering
\includegraphics[scale=0.24]{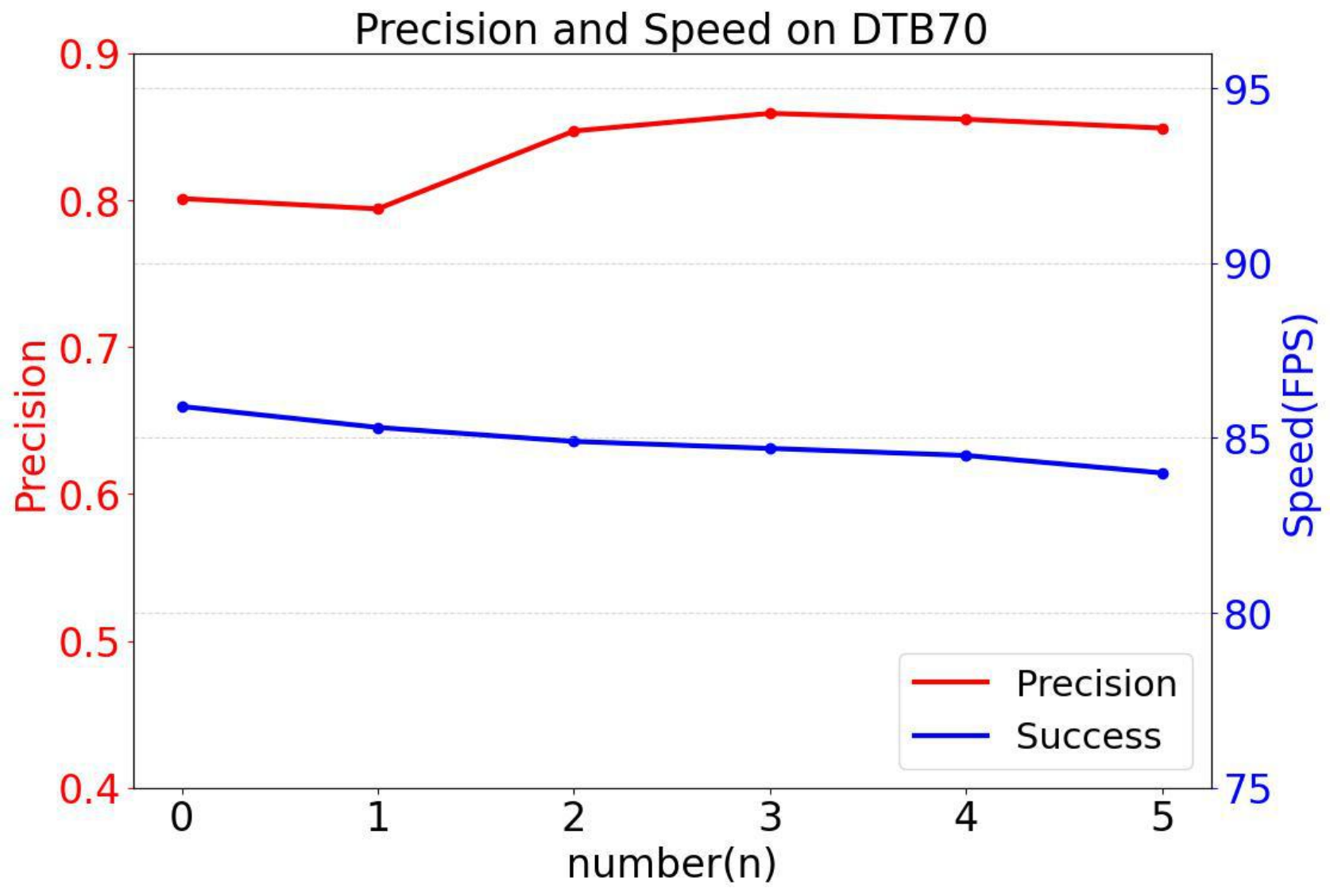}  
\caption
{Precision and Speed of different number of previous frames used in temporal correlation. 
}  
\label{tem_n}    
\end{figure}

\noindent\textbf{Discussion on Mutual Transformer.}
The structure of our proposed mutual transformer is based on the raw transformer, we set up a set of experiments on DTB70 to verify the effect of the part of structure, i.e., encoder and filter.

\textit{Effect of Encoder.} 
Decoder of mutual transformer is the main module to model temporal contexts. While as for encoder, we perform ablation experiments to show its effect, and the results are shown in the fifth and sixth lines of Table \ref{ablation}. The model without encoder decreases the precision and success rate by $6.7\%$ and $7.9\%$ respectively. The results show that the encoder can concentrate on the feature of object, which is helpful for the temporal modeling in mutual transformer.

\textit{Effect of Filter.}
In the process of tracking, the tracker may experience failure and locate the wrong object, which is a distracting information. 
Adding the successive temporal contextual information without filtering out the feeble information will make the tracker confused. As a result, the tracking performance is seriously impaired without adding our filter module in mutual transformer. As shown in the fourth and sixth lines of Table \ref{ablation}, the filter module can adaptively select valid contexts and significantly raise the performance by $3.9\%$ in precision and $5.2\%$ in success rate, demonstrating that filter can effectively remove distracting information.


\begin{figure}[t]
\centering
\includegraphics[scale=0.24]{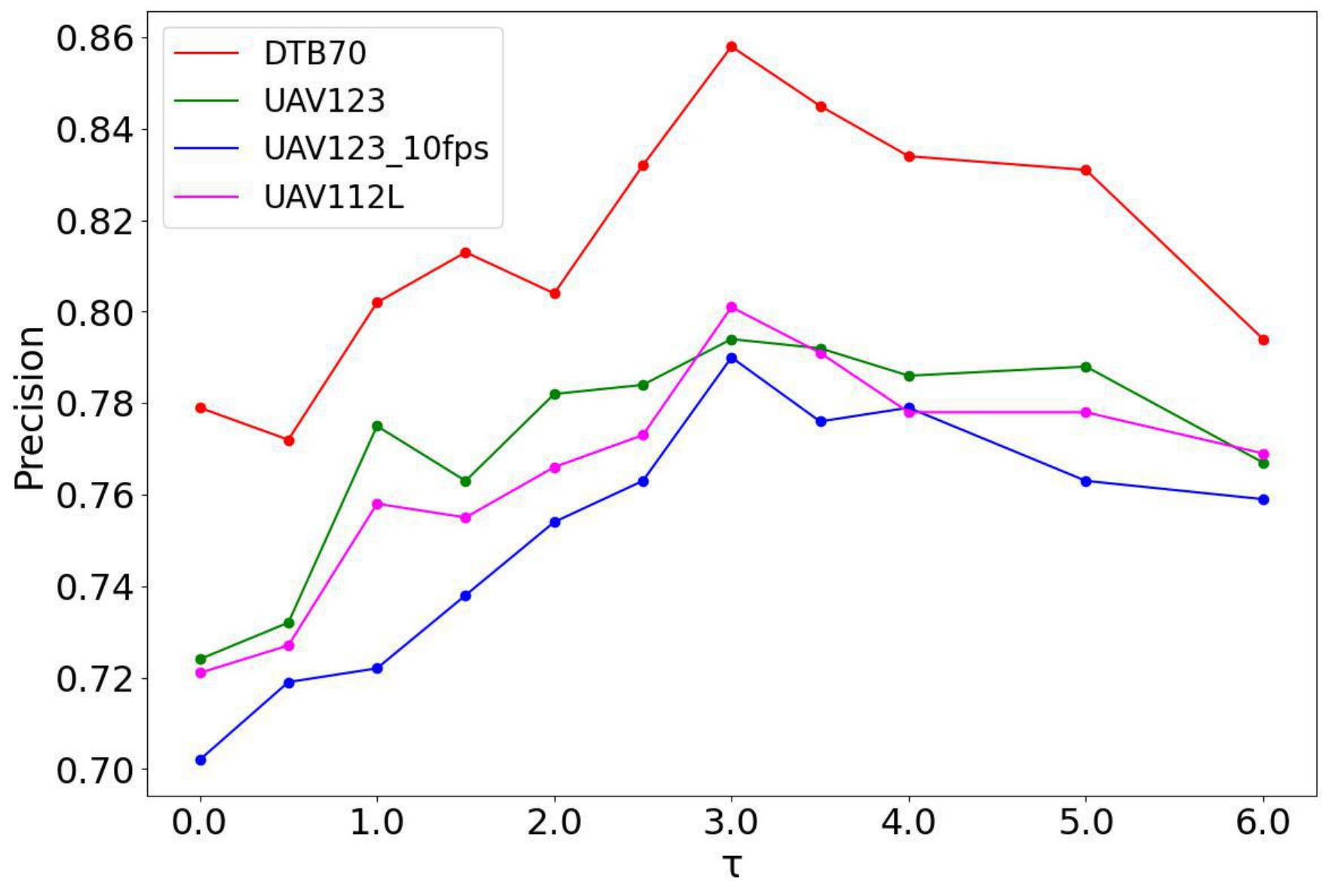}  
\caption
{Analysis on threshold $\tau$. The experiments are conducted on four UAV benchmarks used different threshold $\tau$. We finally choose $\tau=3.0$.
}  
\label{threshold}    
\end{figure}

\noindent\textbf{Discussion on Temporal correlation.} 
\label{n}
In order to improve the performance of our tracker, comprehensive ablation experiments on parameter settings are shown in this subsection.

\textit{Impact of frame length.}
\label{frame length}
The number of previous frames in MTF influences the tracking accuracy as shown in Fig. \ref{tem_n}. We adopt the general depth-wise correlation in the model with $n=0$. Adopting $n=1$ directly lowers the performance of $n=0$ by about $0.7\%$ in precision, which proves that the separate historical frame hardly performs well in temporal correlation. The efficacy of the module begins to manifest significantly from $n=2$ onwards, and the performance peaks at $n=3$ on the DTB70. Beyond this point, further increasing the number of historical frames leads to
a slight decline in accuracy. Crucially, we also observed a trade-off between accuracy and tracking speed. As the number of historical frames increases, the computational load rises, resulting in a reduced tracking speed. Thus, based on these comprehensive experimental findings, we have determined that $n=3$ represents the optimal balance between tracking accuracy and speed. 

As the visualization described in Section \ref{visual}, the first line of Fig. \ref{vis} (b) represents the similarity map produced by standard depth-wise correlation operation, while the second and third lines represent the similarity map produced by our temporal correlation with different historical sequences length. The tracker with $n=3$ concentrates more attention on object and locates the object continuously, while the tracker with $n=1$ concentrates little attention on object and fails quickly.

\textit{Impact of threshold $\tau$.}
\label{thr}
The tracking results are not always correct, taking every previous feature map into account is not reasonable, hence we need to choose appropriate previous feature map. We conduct this experiment to select proper threshold $\tau$ to build more comprehensive temporal contexts. To determine its value, we compare the precision by using different sizes of thresholds on all four benchmarks. Fig. \ref{threshold} shows that $\tau = 3.0$ delivers the best tracking performance.

\begin{table}[t]
\renewcommand\arraystretch{1.2}
\setlength\tabcolsep{6pt}
\caption{Performance comparison of representative DCF and Siamese trackers, in terms of Precision Rate (PR), Normalized Precision Rate (NPR), and Area Under the Curve (AUC), on OTB2015 and Lasot. The best results are shown in \textbf{bold}.\label{general}}
\centering
\begin{tabular}{l| c c c | c c c}
\toprule
\multirow{2}{*}{Method} & \multicolumn{3}{c|}{Lasot} & \multicolumn{3}{c}{OTB2015} \\
\cline{2-7}
~ & PR & NPR & AUC & PR & NPR & AUC \\
\midrule
ECO\cite{danelljan2017eco} & 0.324 & 0.338 & 0.301 & 0.784 &0.592&0.584\\
ATOM\cite{ATOM} & 0.505 & 0.576 & 0.514 & 0.817 & 0.758 & 0.602 \\
DaSiamRPN\cite{DaSiamRPN} & 0.413 & 0.496 & 0.415 & 0.765 &0.577 &0.552\\ Ocean\cite{Ocean} & 0.566 & 0.651 &0.560 &0.785 & 0.707 &0.570 \\
SiamAPN++\cite{SiamAPN++} & 0.547 & 0.638 & 0.559 & 0.824 & 0.784 &0.634 \\
TCTrack\cite{TCTrack} & 0.422 & 0.494 &0.412 &0.825 &0.775 & 0.615 \\ 
\midrule
\textbf{MT-Track (ours)} & \textbf{0.574} & \textbf{0.663} & \textbf{0.569} & \textbf{0.833} & \textbf{0.792} & \textbf{0.642} \\
\bottomrule
\end{tabular}
\end{table}


\begin{table}[t]
\renewcommand{\arraystretch}{1.2}
\setlength{\tabcolsep}{20.0pt}
\caption{The overall average accuracy(\%) of state-of-the-art trackers on Anti-UAV test and validation sets. The best results are shown in \textbf{bold}.\label{anti-uav}}
\centering
\begin{tabular}{l| c c }
\toprule
\multirow{2}{*}{Method} & \multicolumn{2}{c}{Accuracy}\\
\cline{2-3}
  &test & validation \\
\midrule
ARCF\cite{9009064} & 40.55 & 43.82 \\SiamDW\cite{SiamDW} & 38.01 & 49.46 \\
BACF\cite{kiani2017learning} & 36.78 & 43.16 \\ DaSiamRPN\cite{DaSiamRPN} & 39.91 & 44.64 \\
DSST\cite{7569092} & 35.18 & 40.54 \\ SaimRPN++\cite{SiamRPN++} & 43.01 & 48.60 \\
AutoTrack\cite{li2020autotrack} & 38.70 & 49.49 \\ SiamCAR\cite{SiameseCAR} & 47.82 & 56.70 \\
SiamFC\cite{SiamFC} & 36.97 & 49.34 \\ SiamBAN\cite{siamban} & 40.86 & 43.60 \\
SiamMask\cite{Wang2018SiamMask} & 37.44 & 44.34 \\ Hift\cite{Hift} & 37.87 & 47.41 \\
TCTrack\cite{TCTrack} & 52.14 & 57.29 \\ 
PrDimp\cite{PrDimp} & 56.50 & 64.54 \\ 
\midrule
\textbf{MT-Track (ours)} & \textbf{57.92} & \textbf{66.62} \\
\bottomrule
\end{tabular}
\end{table}

\begin{figure}[tp]
\centering
\includegraphics[scale=0.15]{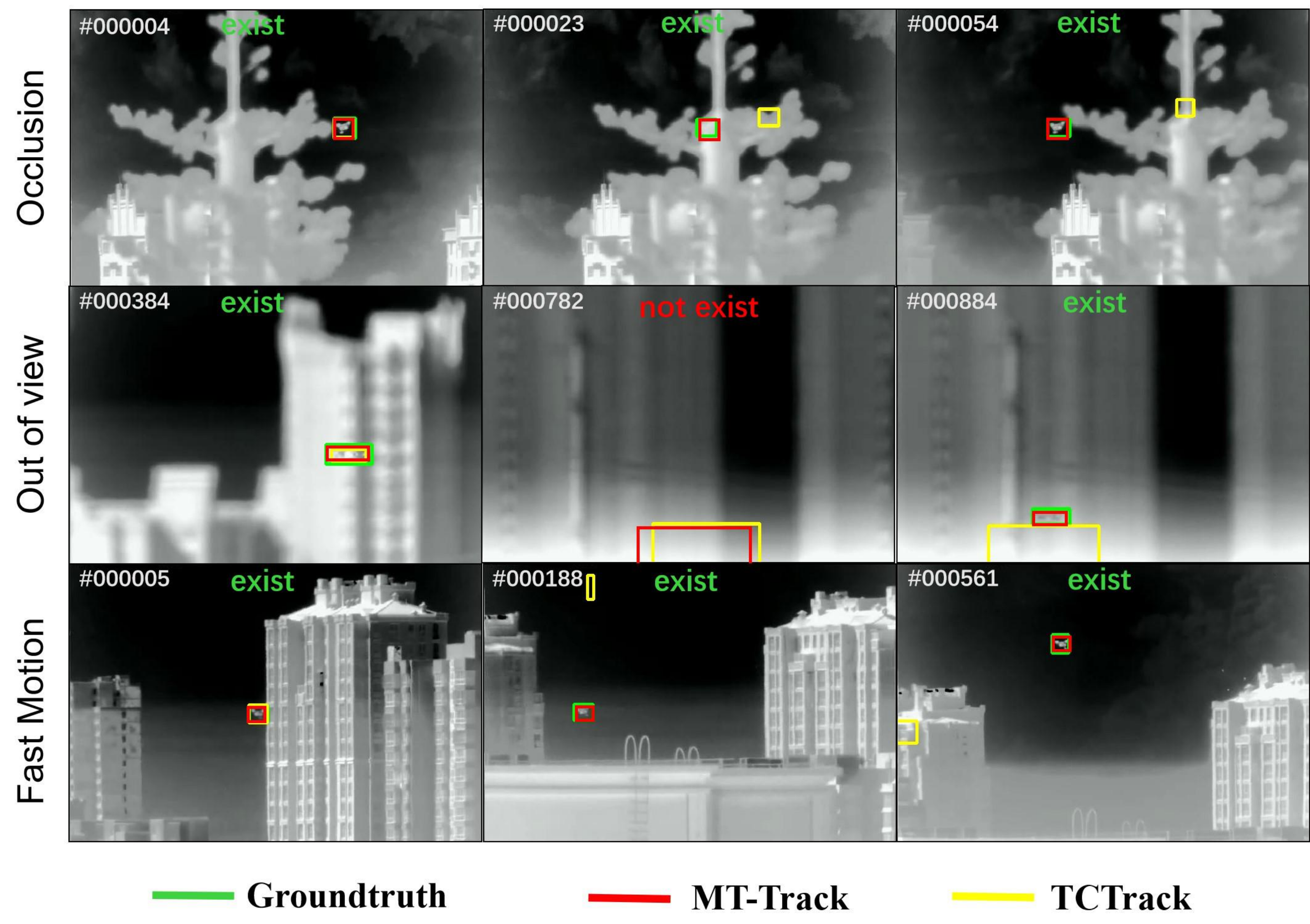}  
\caption
{Qualitative comparison of MT-Track with TCTrack in handling different challenging scenarios in Anti-UAV.}  
\label{antivis}    
\end{figure}
\subsection{Further experiments}
\noindent\textbf{Extension on general object tracking.}
To further evaluate the proposed approach on general object tracking datasets, we conduct experiments on Lasot\cite{lasot} and OTB2015\cite{otb}. We compare our method with methods applied to UAV platforms, and the results are presented in Table \ref{general}. 

\textit{Lasot.}
Lasot\cite{lasot} contains over an average of 2500 densely annotated and high-quality frames for the 280 testing videos, widely used for long-term tracking. Compared with other UAV tracking frameworks, our approach exhibits strong competitive results, and exceeds the second best tracker Ocean by $0.8\%$ (PR),  $1.2\%$ (NPR), and $0.9\%$ (AUC).
This verifies that our temporal modeling module can maintain good and stable performance in the face of challenging long-term tracking tasks.

\textit{OTB2015.}
OTB2015\cite{otb} is one of the most classical benchmarks for object tracking and provides an impartial test of robustness for all trackers. We conduct experiments on OTB2015\cite{otb} which contains 100 videos for tracker performance evaluation. In the existing trackers, our MT-Track yields the best precision rate (0.833), normalized precision rate (0.792), and area under the curve (0.642), which shows that our method has competitive performance in general object tracking.

\noindent\textbf{Extension on Anti-UAV tracking.}
\label{anti}
To verify the performance of MT-Track on the task of tracking UAVs, we further evaluate the tracking performance on Anti-UAV \cite{Anti}. The dataset consists of 318 high-quality RGB-T video pairs, each containing an RGB video and a thermal video. In our evaluation, we conduct the experiment on the thermal infrared sequences, including 160 train sequences, 91 test sequences, and 67 validation sequences. Average accuracy indicates the average overlap ratio between the predicted bounding boxes and ground-truth bounding boxes of all sequences.

We introduce 14 SoTA trackers for comparison, including ARCF\cite{9009064}, BACF\cite{kiani2017learning},  DSST\cite{7569092}, AutoTrack\cite{li2020autotrack}, DaSiamRPN\cite{DaSiamRPN}, SiamRPN++\cite{SiamRPN++}, SiamBAN\cite{siamban}, SiamCAR\cite{SiameseCAR}, SiamFC\cite{SiamFC}, SiamMask\cite{Wang2018SiamMask}, SiamDW\cite{SiamDW}, Hift\cite{Hift}, TCTrack\cite{TCTrack} and PrDimp\cite{PrDimp}.
As shown in Table \ref{anti-uav}, MT-Track performs best in both test and validation sets, achieving the accuracy of $57.92\%$ and $66.62\%$ in test and validation sets, respectively, which proves that our method can be extensively used. In addition, Fig. \ref{antivis} also clearly illustrates that MT-Track maintains leading performance under fast motion, occlusion and out of view.

\begin{figure}[tp]
\centering
\includegraphics[scale=0.31]{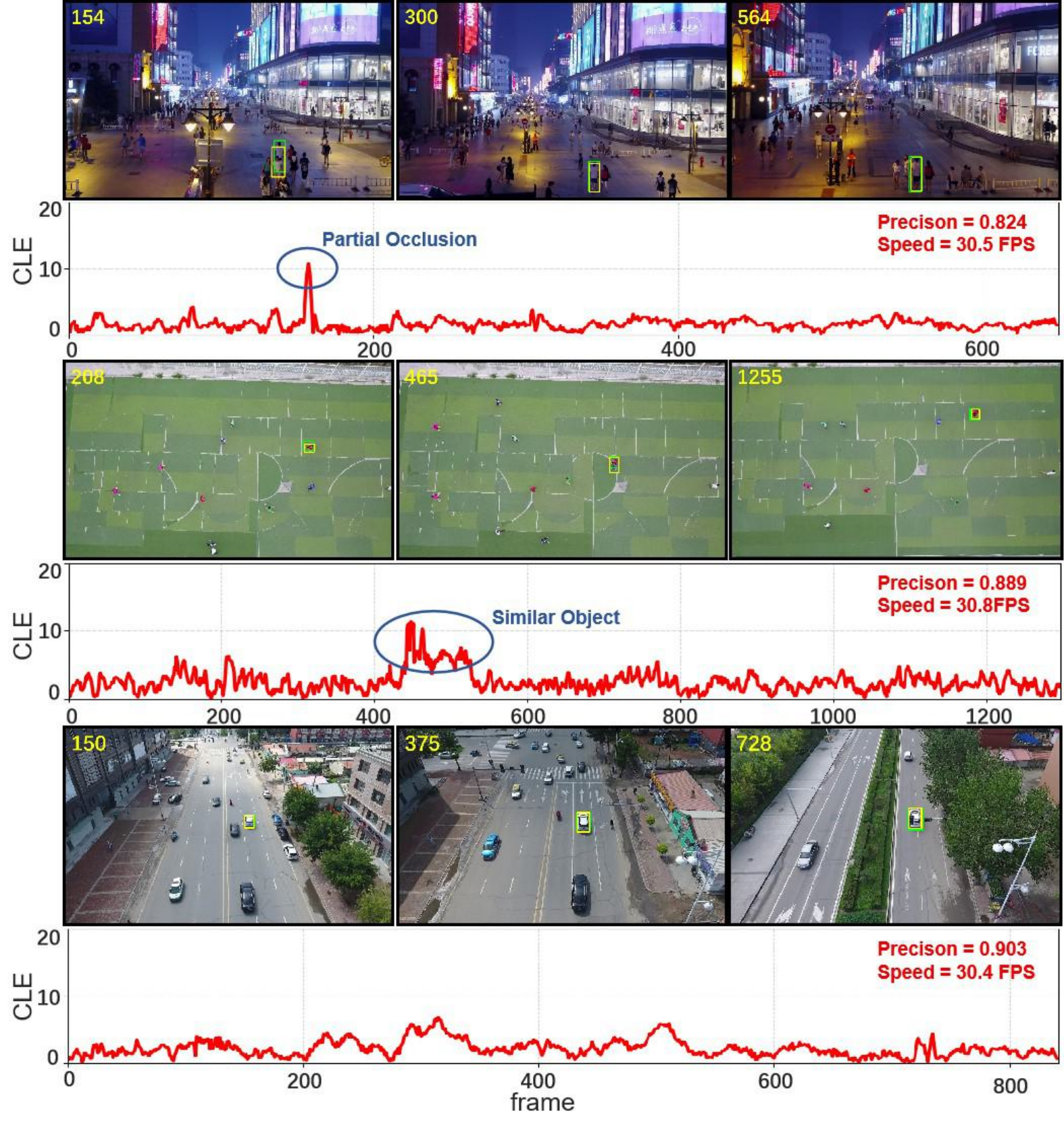}  
\caption
{Visualization of real-world tests on the embedded device. CLE represents the center location error, and the pixel distance threshold is 20. The ground truth and tracking results are marked with \textcolor{green}{green} and \textcolor{yellow}{yellow} boxes.}  
\label{real-world}    
\end{figure}

\noindent\textbf{Onboard Test.}
To validate our tracker’s practicality and robustness, we extend our real-world test, deploying on the NVIDIA Jetson AGX Xavier. This platform is well-suited for UAV applications due to its compactness and computational efficiency.
The VisDrone-SOT dataset \cite{du2019visdrone}, encompassing diverse UAV scenes, serves as our testing scenarios. Fig. \ref{real-world} presents tests in various real-world scenarios, highlighting our tracker’s proficiency in handling partial occlusion, camera motion, and small objects. The evaluation metric used is the Center Location Error (CLE), providing a precise measure of tracking accuracy. A CLE score below 20 pixels is considered as a successful tracking result. Our findings demonstrate that MT-Track maintains robust tracking performance and achieves real-time speeds exceeding 30 FPS in these challenging conditions.

\begin{figure}[t]
\centering
\includegraphics[scale=0.39]{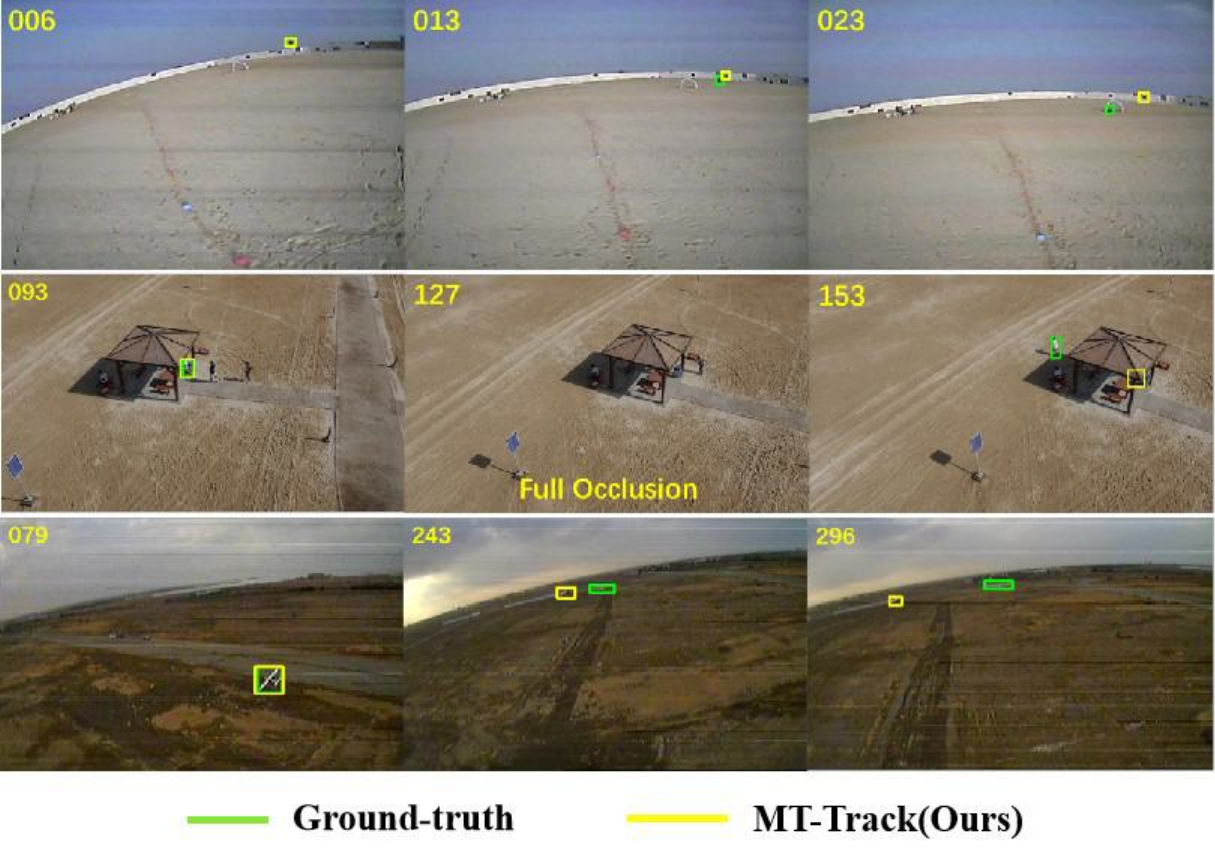}  
\caption
{Failed Cases. MT-Track fails to localize the target when the target is extremely small, appears to be fully occluded or is very similar to the background.}  
\label{fail}    
\end{figure}

\section{Limitations and Future Work}
\subsection{Limitations}
Failure cases are illustrated in Fig. \ref{fail}. Our tracker’s local search strategy does not include a global detector, which leads to difficulties in reacquiring the target post-complete occlusion (the second row). This limitation hinders the continuity of temporal modeling and effective track reconstruction when the target reappears.
In addition, while our tracker shows advancements in handling low-resolution and cluttered backgrounds, its performance is constrained when dealing with extremely small or camouflaged targets (the first and third rows). This is primarily due to the use of a shallow feature extraction network, which, though beneficial for real-time processing on UAV platforms, provides a limited feature set for accurate foreground-background differentiation and temporal modeling.

\subsection{Future Work}
The global detector is planned to be incorporated into our tracking framework. This enhancement aims to augment the robustness of our tracker, particularly in scenarios involving full occlusion and complex environments. Moreover, future research will continue to explore deep feature extraction networks that are not only efficient for real-time UAV tracking but also versatile for broader applications in computer vision. These networks will aim to provide richer feature sets for improved target differentiation and temporal analysis.

\section{Conclusion}
In this work, we propose a concise and efficient multi-step temporal modeling module, which fully extracts the temporal contexts of video sequences. 
First, we propose a temporal correlation module to generate more accurate correlation maps by updating template feature. 
Next, we propose a mutual transformer module to model temporal knowledge and refine the correlation maps. 
The proposed MT-Track gets rid of hyper-parameters sensitive post-processing, leading to a simple inference pipeline and enabling efficient temporal modeling.
Extensive experiments have shown that MT-Track performs well on four authoritative aerial tracking benchmarks while running at 84.7 FPS on a single GPU and 30.4 FPS on embedded device. We hope that our framework can attract more attention on aerial and even general tracking with temporal information.

{
\bibliographystyle{IEEEtran}
\bibliography{IEEEabrv,egbib}

\begin{thebibliography}{100}

\bibitem{SiamFC}
Luca Bertinetto, Jack Valmadre, Joao~F Henriques, Andrea Vedaldi, and Philip~HS
  Torr.
\newblock Fully-convolutional siamese networks for object tracking.
\newblock In {\em European conference on computer vision}, pages 850--865.
  Springer, 2016.

\bibitem{08}
Goutam Bhat, Martin Danelljan, Luc~Van Gool, and Radu Timofte.
\newblock Learning discriminative model prediction for tracking.
\newblock In {\em Proceedings of the IEEE/CVF international conference on
  computer vision}, pages 6182--6191, 2019.

\bibitem{HCAT}
Philippe Blatter, Menelaos Kanakis, Martin Danelljan, and Luc~Van Gool.
\newblock Efficient visual tracking with exemplar transformers.
\newblock {\em CoRR}, abs/2112.09686, 2021.

\bibitem{blatter2023efficient}
Philippe Blatter, Menelaos Kanakis, Martin Danelljan, and Luc Van~Gool.
\newblock Efficient visual tracking with exemplar transformers.
\newblock In {\em Proceedings of the IEEE/CVF Winter Conference on Applications
  of Computer Vision}, pages 1571--1581, 2023.

\bibitem{agri}
Weihao Bo, Jie Liu, Xijian Fan, Tardi Tjahjadi, Qiaolin Ye, and Liyong Fu.
\newblock Basnet: Burned area segmentation network for real-time detection of
  damage maps in remote sensing images.
\newblock {\em IEEE Transactions on Geoscience and Remote Sensing}, 60:1--13,
  2022.

\bibitem{Hift}
Ziang Cao, Changhong Fu, Junjie Ye, Bowen Li, and Yiming Li.
\newblock Hift: Hierarchical feature transformer for aerial tracking.
\newblock In {\em Proceedings of the IEEE/CVF International Conference on
  Computer Vision}, pages 15457--15466, 2021.

\bibitem{SiamAPN++}
Ziang Cao, Changhong Fu, Junjie Ye, Bowen Li, and Yiming Li.
\newblock Siamapn++: Siamese attentional aggregation network for real-time uav
  tracking.
\newblock In {\em 2021 IEEE/RSJ International Conference on Intelligent Robots
  and Systems (IROS)}, pages 3086--3092. IEEE, 2021.

\bibitem{TCTrack}
Ziang Cao, Ziyuan Huang, Liang Pan, Shiwei Zhang, Ziwei Liu, and Changhong Fu.
\newblock Tctrack: Temporal contexts for aerial tracking.
\newblock In {\em Proceedings of the IEEE/CVF Conference on Computer Vision and
  Pattern Recognition}, pages 14798--14808, 2022.

\bibitem{DETR}
Nicolas Carion, Francisco Massa, Gabriel Synnaeve, Nicolas Usunier, Alexander
  Kirillov, and Sergey Zagoruyko.
\newblock End-to-end object detection with transformers.
\newblock In {\em European conference on computer vision}, pages 213--229.
  Springer, 2020.

\bibitem{10144283}
Guanlin Chen, Pengfei Zhu, Bing Cao, Xing Wang, and Qinghua Hu.
\newblock Cross-drone transformer network for robust single object tracking.
\newblock {\em IEEE Transactions on Circuits and Systems for Video Technology},
  33(9):4552--4563, 2023.

\bibitem{06}
Peng Chen, Yuanjie Dang, Ronghua Liang, Wei Zhu, and Xiaofei He.
\newblock Real-time object tracking on a drone with multi-inertial sensing
  data.
\newblock {\em IEEE Transactions on Intelligent Transportation Systems},
  19(1):131--139, 2017.

\bibitem{Transformertracking}
Xin Chen, Bin Yan, Jiawen Zhu, Dong Wang, Xiaoyun Yang, and Huchuan Lu.
\newblock Transformer tracking.
\newblock In {\em Proceedings of the IEEE/CVF conference on computer vision and
  pattern recognition}, pages 8126--8135, 2021.

\bibitem{siamban}
Zedu Chen, Bineng Zhong, Guorong Li, Shengping Zhang, and Rongrong Ji.
\newblock Siamese box adaptive network for visual tracking.
\newblock In {\em Proceedings of the IEEE Conference on Computer Vision and
  Pattern Recognition}, 2020.

\bibitem{superdimp}
Seokeon Choi, Junhyun Lee, Yunsung Lee, and Alexander Hauptmann.
\newblock Robust long-term object tracking via improved discriminative model
  prediction.
\newblock In {\em Proceedings of the European Conference on Computer Vision
  (ECCV)}, pages 0--0, 2020.

\bibitem{PrDimp}
M.~Danelljan, L~Van Gool, and R.~Timofte.
\newblock Probabilistic regression for visual tracking.
\newblock 2020.

\bibitem{ATOM}
Martin Danelljan, Goutam Bhat, Fahad~Shahbaz Khan, and Michael Felsberg.
\newblock Atom: Accurate tracking by overlap maximization.
\newblock In {\em 2019 IEEE/CVF Conference on Computer Vision and Pattern
  Recognition (CVPR)}, pages 4655--4664, 2019.

\bibitem{8953466}
Martin Danelljan, Goutam Bhat, Fahad~Shahbaz Khan, and Michael Felsberg.
\newblock Atom: Accurate tracking by overlap maximization.
\newblock In {\em 2019 IEEE/CVF Conference on Computer Vision and Pattern
  Recognition (CVPR)}, pages 4655--4664, 2019.

\bibitem{danelljan2017eco}
Martin Danelljan, Goutam Bhat, Fahad Shahbaz~Khan, and Michael Felsberg.
\newblock Eco: Efficient convolution operators for tracking.
\newblock In {\em Proceedings of the IEEE conference on computer vision and
  pattern recognition}, pages 6638--6646, 2017.

\bibitem{09}
Martin Danelljan, Luc~Van Gool, and Radu Timofte.
\newblock Probabilistic regression for visual tracking.
\newblock In {\em Proceedings of the IEEE/CVF conference on computer vision and
  pattern recognition}, pages 7183--7192, 2020.

\bibitem{7569092}
Martin Danelljan, Gustav Häger, Fahad~Shahbaz Khan, and Felsberg.
\newblock Discriminative scale space tracking.
\newblock {\em IEEE Transactions on Pattern Analysis and Machine Intelligence},
  39(8):1561--1575, 2017.

\bibitem{7410847}
Martin Danelljan, Gustav Häger, Fahad~Shahbaz Khan, and Michael Felsberg.
\newblock Learning spatially regularized correlation filters for visual
  tracking.
\newblock In {\em 2015 IEEE International Conference on Computer Vision
  (ICCV)}, pages 4310--4318, 2015.

\bibitem{danelljan2016beyond}
Martin Danelljan, Andreas Robinson, Fahad Shahbaz~Khan, and Michael Felsberg.
\newblock Beyond correlation filters: Learning continuous convolution operators
  for visual tracking.
\newblock In {\em European conference on computer vision}, pages 472--488.
  Springer, 2016.

\bibitem{9447987}
Chenwei Deng, Shuangcheng He, Yuqi Han, and Boya Zhao.
\newblock Learning dynamic spatial-temporal regularization for uav object
  tracking.
\newblock {\em IEEE Signal Processing Letters}, 28:1230--1234, 2021.

\bibitem{VIT}
Alexey Dosovitskiy, Lucas Beyer, Alexander Kolesnikov, Dirk Weissenborn,
  Xiaohua Zhai, Thomas Unterthiner, Mostafa Dehghani, Matthias Minderer, Georg
  Heigold, Sylvain Gelly, et~al.
\newblock An image is worth 16x16 words: Transformers for image recognition at
  scale.
\newblock {\em arXiv preprint arXiv:2010.11929}, 2020.

\bibitem{du2019visdrone}
Dawei Du, Pengfei Zhu, Longyin Wen, Xiao Bian, Haibin Ling, Qinghua Hu, Jiayu
  Zheng, Tao Peng, Xinyao Wang, Yue Zhang, et~al.
\newblock Visdrone-sot2019: The vision meets drone single object tracking
  challenge results.
\newblock In {\em Proceedings of the IEEE/CVF International Conference on
  Computer Vision Workshops}, pages 0--0, 2019.

\bibitem{lasot}
Heng Fan, Liting Lin, Fan Yang, Peng Chu, Ge~Deng, Sijia Yu, Hexin Bai, Yong
  Xu, Chunyuan Liao, and Haibin Ling.
\newblock Lasot: A high-quality benchmark for large-scale single object
  tracking.
\newblock In {\em Proceedings of the IEEE/CVF Conference on Computer Vision and
  Pattern Recognition (CVPR)}, June 2019.

\bibitem{9064815}
Jiaqing Fan, Huihui Song, Kaihua Zhang, Kang Yang, and Qingshan Liu.
\newblock Feature alignment and aggregation siamese networks for fast visual
  tracking.
\newblock {\em IEEE Transactions on Circuits and Systems for Video Technology},
  31(4):1296--1307, 2021.

\bibitem{SiamAPN}
Changhong Fu, Ziang Cao, Yiming Li, Junjie Ye, and Chen Feng.
\newblock Siamese anchor proposal network for high-speed aerial tracking.
\newblock In {\em 2021 IEEE International Conference on Robotics and Automation
  (ICRA)}, pages 510--516. IEEE, 2021.

\bibitem{9477413}
Changhong Fu, Ziang Cao, Yiming Li, Junjie Ye, and Chen Feng.
\newblock Onboard real-time aerial tracking with efficient siamese anchor
  proposal network.
\newblock {\em IEEE Transactions on Geoscience and Remote Sensing}, 60:1--13,
  2022.

\bibitem{fu2021correlation}
Changhong Fu, Bowen Li, Fangqiang Ding, Fuling Lin, and Geng Lu.
\newblock Correlation filters for unmanned aerial vehicle-based aerial
  tracking: A review and experimental evaluation.
\newblock {\em IEEE Geoscience and Remote Sensing Magazine}, 10(1):125--160,
  2021.

\bibitem{STMTrack}
Zhihong Fu, Qingjie Liu, Zehua Fu, and Yunhong Wang.
\newblock Stmtrack: Template-free visual tracking with space-time memory
  networks.
\newblock In {\em Proceedings of the IEEE/CVF Conference on Computer Vision and
  Pattern Recognition}, pages 13774--13783, 2021.

\bibitem{graph}
Junyu Gao, Tianzhu Zhang, and Changsheng Xu.
\newblock Graph convolutional tracking.
\newblock pages 4649--4659, 2019.

\bibitem{guo2023vit}
Chuangye Guo, Kang Liu, Donghu Deng, and Xuelong Li.
\newblock Vit spatio-temporal feature fusion for aerial object tracking.
\newblock {\em IEEE Transactions on Circuits and Systems for Video Technology},
  2023.

\bibitem{siamgat}
Dongyan Guo, Yanyan Shao, Ying Cui, Zhenhua Wang, Liyan Zhang, and Chunhua
  Shen.
\newblock Graph attention tracking.
\newblock In {\em The IEEE Conference on Computer Vision and Pattern
  Recognition (CVPR)}, June 2021.

\bibitem{SiameseCAR}
Dongyan Guo, Jun Wang, Ying Cui, Zhenhua Wang, and Shengyong Chen.
\newblock Siamcar: Siamese fully convolutional classification and regression
  for visual tracking.
\newblock In {\em Proceedings of the IEEE/CVF conference on computer vision and
  pattern recognition}, pages 6269--6277, 2020.

\bibitem{DSiam}
Qing Guo, Wei Feng, Ce~Zhou, Rui Huang, Liang Wan, and Song Wang.
\newblock Learning dynamic siamese network for visual object tracking.
\newblock In {\em Proceedings of the IEEE international conference on computer
  vision}, pages 1763--1771, 2017.

\bibitem{9613749}
Guang Han, Jinpeng Su, Yaoming Liu, Yuqiu Zhao, and Sam Kwong.
\newblock Multi-stage visual tracking with siamese anchor-free proposal
  network.
\newblock {\em IEEE Transactions on Multimedia}, 25:430--442, 2023.

\bibitem{han2021connection}
Qi~Han, Zejia Fan, Qi~Dai, Lei Sun, Ming-Ming Cheng, Jiaying Liu, and Jingdong
  Wang.
\newblock On the connection between local attention and dynamic depth-wise
  convolution.
\newblock In {\em International Conference on Learning Representations}, 2021.

\bibitem{Han}
Yuqi Han, Chenwei Deng, Boya Zhao, and Baojun Zhao.
\newblock Spatial-temporal context-aware tracking.
\newblock {\em IEEE signal processing letters}, 26(3):500--504, 1 2019.

\bibitem{han2022comprehensive}
Yuqi Han, Huaping Liu, Yufeng Wang, and Chunlei Liu.
\newblock A comprehensive review for typical applications based upon unmanned
  aerial vehicle platform.
\newblock {\em IEEE Journal of Selected Topics in Applied Earth Observations
  and Remote Sensing}, 15:9654--9666, 2022.

\bibitem{2020Boundary}
Yuqi Han, Hongshuo Wang, Zengshuo Zhang, and Zheng Wang.
\newblock Boundary-aware vehicle tracking upon uav.
\newblock {\em Electronics Letters}, 2020.

\bibitem{6870486}
João~F. Henriques, Rui Caseiro, Pedro Martins, and Jorge Batista.
\newblock High-speed tracking with kernelized correlation filters.
\newblock {\em IEEE Transactions on Pattern Analysis and Machine Intelligence},
  37(3):583--596, 2015.

\bibitem{SE-Net}
Jie Hu, Li~Shen, Samuel Albanie, Gang Sun, and Enhua Wu.
\newblock Squeeze-and-excitation networks.
\newblock Number {arXiv}:1709.01507.

\bibitem{globaltrack}
Lianghua Huang, Xin Zhao, and Kaiqi Huang.
\newblock Globaltrack: {A} simple and strong baseline for long-term tracking.
\newblock {\em CoRR}, abs/1912.08531, 2019.

\bibitem{got}
Lianghua Huang, Xin Zhao, and Kaiqi Huang.
\newblock Got-10k: A large high-diversity benchmark for generic object tracking
  in the wild.
\newblock {\em IEEE Transactions on Pattern Analysis and Machine Intelligence},
  43(5):1562--1577, 2019.

\bibitem{huang2019learning}
Ziyuan Huang, Changhong Fu, Yiming Li, Fuling Lin, and Peng Lu.
\newblock Learning aberrance repressed correlation filters for real-time uav
  tracking.
\newblock In {\em Proceedings of the IEEE/CVF International Conference on
  Computer Vision}, pages 2891--2900, 2019.

\bibitem{9009064}
Ziyuan Huang, Changhong Fu, Yiming Li, Fuling Lin, and Peng Lu.
\newblock Learning aberrance repressed correlation filters for real-time uav
  tracking.
\newblock In {\em 2019 IEEE/CVF International Conference on Computer Vision
  (ICCV)}, pages 2891--2900, 2019.

\bibitem{tada}
Ziyuan Huang, Shiwei Zhang, Liang Pan, Zhiwu Qing, Mingqian Tang, Ziwei Liu,
  and Marcelo~H Ang~Jr.
\newblock Tada! temporally-adaptive convolutions for video understanding.
\newblock {\em arXiv preprint arXiv:2110.06178}, 2021.

\bibitem{Anti}
Nan Jiang, Kuiran Wang, Xiaoke Peng, Xuehui Yu, Qiang Wang, Junliang Xing,
  Guorong Li, Jian Zhao, Guodong Guo, and Zhenjun Han.
\newblock Anti-uav: {A} large multi-modal benchmark for {UAV} tracking.
\newblock {\em CoRR}, abs/2101.08466, 2021.

\bibitem{kiani2017learning}
Hamed Kiani~Galoogahi, Ashton Fagg, and Simon Lucey.
\newblock Learning background-aware correlation filters for visual tracking.
\newblock In {\em Proceedings of the IEEE international conference on computer
  vision}, pages 1135--1143, 2017.

\bibitem{LTDSE}
M.~Kristan, A.~Berg, L.~Zheng, L.~Rout, and L.~Zhou.
\newblock The seventh visual object tracking vot2019 challenge results.
\newblock In {\em 2019 IEEE/CVF International Conference on Computer Vision
  Workshop (ICCVW)}, 2019.

\bibitem{alxnet}
Alex Krizhevsky, Ilya Sutskever, and Geoffrey~E Hinton.
\newblock Imagenet classification with deep convolutional neural networks.
\newblock {\em Communications of the ACM}, 60(6):84--90, 2017.

\bibitem{03}
Kuan-Hui Lee and Jenq-Neng Hwang.
\newblock On-road pedestrian tracking across multiple driving recorders.
\newblock {\em IEEE Transactions on Multimedia}, 17(9):1429--1438, 2015.

\bibitem{SiamRPN++}
Bo~Li, Wei Wu, Qiang Wang, Fangyi Zhang, Junliang Xing, and Junjie Yan.
\newblock Siamrpn++: Evolution of siamese visual tracking with very deep
  networks.
\newblock In {\em Proceedings of the IEEE/CVF Conference on Computer Vision and
  Pattern Recognition}, pages 4282--4291, 2019.

\bibitem{SiamRPN}
Bo~Li, Junjie Yan, Wei Wu, Zheng Zhu, and Xiaolin Hu.
\newblock High performance visual tracking with siamese region proposal
  network.
\newblock In {\em Proceedings of the IEEE conference on computer vision and
  pattern recognition}, pages 8971--8980, 2018.

\bibitem{dtb}
Siyi Li and Dit-Yan Yeung.
\newblock Visual object tracking for unmanned aerial vehicles: A benchmark and
  new motion models.
\newblock In {\em Thirty-first AAAI conference on artificial intelligence},
  2017.

\bibitem{li2019target}
Xin Li, Chao Ma, Baoyuan Wu, Zhenyu He, and Ming-Hsuan Yang.
\newblock Target-aware deep tracking.
\newblock In {\em Proceedings of the IEEE/CVF conference on computer vision and
  pattern recognition}, pages 1369--1378, 2019.

\bibitem{li2020autotrack}
Yiming Li, Changhong Fu, Fangqiang Ding, Ziyuan Huang, and Geng Lu.
\newblock Autotrack: Towards high-performance visual tracking for uav with
  automatic spatio-temporal regularization.
\newblock In {\em Proceedings of the IEEE/CVF Conference on Computer Vision and
  Pattern Recognition}, pages 11923--11932, 2020.

\bibitem{10.1007/978-3-030-58542-6_26}
Bingyan Liao, Chenye Wang, Yayun Wang, Yaonong Wang, and Jun Yin.
\newblock Pg-net: Pixel to global matching network for visual tracking.
\newblock In Andrea Vedaldi, Horst Bischof, Thomas Brox, and Jan-Michael Frahm,
  editors, {\em Computer Vision -- ECCV 2020}, pages 429--444, Cham, 2020.
  Springer International Publishing.

\bibitem{6}
Fuling Lin, Changhong Fu, Yujie He, Fuyu Guo, and Qian Tang.
\newblock Learning temporary block-based bidirectional incongruity-aware
  correlation filters for efficient uav object tracking.
\newblock {\em IEEE Transactions on Circuits and Systems for Video Technology},
  31(6):2160--2174, 2021.

\bibitem{ill}
Shuang Luo, Huifang Li, Yiqiu Li, Chenglin Shao, Huanfeng Shen, and Liangpei
  Zhang.
\newblock An evolutionary shadow correction network and a benchmark uav dataset
  for remote sensing images.
\newblock {\em IEEE Transactions on Geoscience and Remote Sensing}, 61:1--14,
  2023.

\bibitem{TrackFormer}
Tim Meinhardt, Alexander Kirillov, Laura Leal-Taixe, and Christoph
  Feichtenhofer.
\newblock Trackformer: Multi-object tracking with transformers.
\newblock In {\em Proceedings of the IEEE/CVF Conference on Computer Vision and
  Pattern Recognition}, pages 8844--8854, 2022.

\bibitem{UAV123}
Matthias Mueller, Neil Smith, and Bernard Ghanem.
\newblock A benchmark and simulator for uav tracking.
\newblock In {\em European conference on computer vision}, pages 445--461.
  Springer, 2016.

\bibitem{8953982}
Hamid Rezatofighi, Nathan Tsoi, JunYoung Gwak, Amir Sadeghian, Ian Reid, and
  Silvio Savarese.
\newblock Generalized intersection over union: A metric and a loss for bounding
  box regression.
\newblock In {\em 2019 IEEE/CVF Conference on Computer Vision and Pattern
  Recognition (CVPR)}, pages 658--666, 2019.

\bibitem{VID}
Olga Russakovsky, Jia Deng, Hao Su, Jonathan Krause, Sanjeev Satheesh, Sean Ma,
  Zhiheng Huang, Andrej Karpathy, Aditya Khosla, Michael Bernstein, et~al.
\newblock Imagenet large scale visual recognition challenge.
\newblock {\em International journal of computer vision}, 115(3):211--252,
  2015.

\bibitem{russakovsky2015imagenet}
Olga Russakovsky, Jia Deng, Hao Su, Jonathan Krause, Sanjeev Satheesh, Sean Ma,
  Zhiheng Huang, Andrej Karpathy, Aditya Khosla, Michael Bernstein, et~al.
\newblock Imagenet large scale visual recognition challenge.
\newblock {\em International journal of computer vision}, 115(3):211--252,
  2015.

\bibitem{SE-SiamFC}
Ivan Sosnovik, Artem Moskalev, and Arnold~WM Smeulders.
\newblock Scale equivariance improves siamese tracking.
\newblock In {\em Proceedings of the IEEE/CVF Winter Conference on Applications
  of Computer Vision}, pages 2765--2774, 2021.

\bibitem{forest}
S.~Sudhakar, V.~Vijayakumar, C.~{Sathiya Kumar}, V.~Priya, Logesh Ravi, and
  V.~Subramaniyaswamy.
\newblock Unmanned aerial vehicle (uav) based forest fire detection and
  monitoring for reducing false alarms in forest-fires.
\newblock {\em Computer Communications}, 149:1--16, 2020.

\bibitem{TransTrack}
Peize Sun, Jinkun Cao, Yi~Jiang, Rufeng Zhang, Enze Xie, Zehuan Yuan, Changhu
  Wang, and Ping Luo.
\newblock Transtrack: Multiple object tracking with transformer.
\newblock {\em arXiv preprint arXiv:2012.15460}, 2020.

\bibitem{road}
Yueming Sun, Zhenfeng Shao, Gui Cheng, Xiao Huang, and Zhongyuan Wang.
\newblock Road and car extraction using uav images via efficient dual
  contextual parsing network.
\newblock {\em IEEE Transactions on Geoscience and Remote Sensing}, 60:1--13,
  2022.

\bibitem{7780527}
Ran Tao, Efstratios Gavves, and Arnold W.~M. Smeulders.
\newblock Siamese instance search for tracking.
\newblock In {\em 2016 IEEE Conference on Computer Vision and Pattern
  Recognition (CVPR)}, pages 1420--1429, 2016.

\bibitem{9026819}
Shengjing Tian, Xiuping Liu, Meng Liu, Shuhua Li, and Baocai Yin.
\newblock Siamese tracking network with informative enhanced loss.
\newblock {\em IEEE Transactions on Multimedia}, 23:120--132, 2021.

\bibitem{05}
Wei Tian and Martin Lauer.
\newblock Tracking objects with severe occlusion by adaptive part filter
  modeling-in traffic scenes and beyond.
\newblock {\em IEEE Intelligent Transportation Systems Magazine}, 10(4):60--73,
  2018.

\bibitem{9583266}
Xiaoyang Tian, Jie Shao, Deqiang Ouyang, and Heng~Tao Shen.
\newblock Uav-satellite view synthesis for cross-view geo-localization.
\newblock {\em IEEE Transactions on Circuits and Systems for Video Technology},
  32(7):4804--4815, 2022.

\bibitem{Transformer}
Ashish Vaswani, Noam Shazeer, Niki Parmar, Jakob Uszkoreit, Llion Jones,
  Aidan~N Gomez, {\L}ukasz Kaiser, and Illia Polosukhin.
\newblock Attention is all you need.
\newblock {\em Advances in neural information processing systems}, 30, 2017.

\bibitem{Siamrcnn}
Paul Voigtlaender, Jonathon Luiten, Philip H.~S. Torr, and Bastian Leibe.
\newblock Siam r-cnn: Visual tracking by re-detection.
\newblock {\em Computer Vision and Pattern Recognition}, 2020.

\bibitem{TMT}
Ning Wang, Wengang Zhou, Jie Wang, and Houqiang Li.
\newblock Transformer meets tracker: Exploiting temporal context for robust
  visual tracking.
\newblock In {\em Proceedings of the IEEE/CVF Conference on Computer Vision and
  Pattern Recognition}, pages 1571--1580, 2021.

\bibitem{Wang2018SiamMask}
Qiang Wang, Li~Zhang, Luca Bertinetto, Weiming Hu, and Philip~HS Torr.
\newblock Fast online object tracking and segmentation: A unifying approach.
\newblock {\em arXiv preprint arXiv:1812.05050}, 2018.

\bibitem{8457310}
Qiurui Wang, Chun Yuan, Jingdong Wang, and Wenjun Zeng.
\newblock Learning attentional recurrent neural network for visual tracking.
\newblock {\em IEEE Transactions on Multimedia}, 21(4):930--942, 2019.

\bibitem{otb}
Yi~Wu, Jongwoo Lim, and Ming-Hsuan Yang.
\newblock Object tracking benchmark.
\newblock {\em IEEE Transactions on Pattern Analysis and Machine Intelligence},
  37(9):1834--1848, 2015.

\bibitem{9157777}
Yue Wu, Yinpeng Chen, Lu~Yuan, Zicheng Liu, Lijuan Wang, Hongzhi Li, and Yun
  Fu.
\newblock Rethinking classification and localization for object detection.
\newblock In {\em 2020 IEEE/CVF Conference on Computer Vision and Pattern
  Recognition (CVPR)}, pages 10183--10192, 2020.

\bibitem{9448479}
Mao Xi, Wengang Zhou, Ning Wang, and Houqiang Li.
\newblock Learning temporal-correlated and channel- decorrelated siamese
  networks for visual tracking.
\newblock {\em IEEE Transactions on Multimedia}, 24:2791--2803, 2022.

\bibitem{SiamTPN}
Daitao Xing, Nikolaos Evangeliou, Athanasios Tsoukalas, and Anthony Tzes.
\newblock Siamese transformer pyramid networks for real-time uav tracking.
\newblock In {\em Proceedings of the IEEE/CVF Winter Conference on Applications
  of Computer Vision}, pages 2139--2148, 2022.

\bibitem{01}
Junliang Xing, Haizhou Ai, and Shihong Lao.
\newblock Multiple human tracking based on multi-view upper-body detection and
  discriminative learning.
\newblock In {\em 2010 20th International Conference on Pattern Recognition},
  pages 1698--1701. IEEE, 2010.

\bibitem{xu2016deep}
Lie Xu, Chiu-sing Choy, and Yi-Wen Li.
\newblock Deep sparse rectifier neural networks for speech denoising.
\newblock In {\em 2016 IEEE International Workshop on Acoustic Signal
  Enhancement (IWAENC)}, pages 1--5. IEEE, 2016.

\bibitem{SiamFC++}
Yinda Xu, Zeyu Wang, Zuoxin Li, Ye~Yuan, and Gang Yu.
\newblock Siamfc++: Towards robust and accurate visual tracking with target
  estimation guidelines.
\newblock In {\em Proceedings of the AAAI Conference on Artificial
  Intelligence}, volume~34, pages 12549--12556, 2020.

\bibitem{3}
Wanli Xue, Chao Xu, and Zhiyong Feng.
\newblock Robust visual tracking via multi-scale spatio-temporal context
  learning.
\newblock {\em IEEE Transactions on Circuits and Systems for Video Technology},
  28(10):2849--2860, 2017.

\bibitem{Stark}
Bin Yan, Houwen Peng, Jianlong Fu, Dong Wang, and Huchuan Lu.
\newblock Learning spatio-temporal transformer for visual tracking.
\newblock In {\em Proceedings of the IEEE/CVF International Conference on
  Computer Vision}, pages 10448--10457, 2021.

\bibitem{9577854}
Bin Yan, Xinyu Zhang, Dong Wang, Huchuan Lu, and Xiaoyun Yang.
\newblock Alpha-refine: Boosting tracking performance by precise bounding box
  estimation.
\newblock In {\em 2021 IEEE/CVF Conference on Computer Vision and Pattern
  Recognition (CVPR)}, pages 5285--5294, 2021.

\bibitem{10}
Tianyu Yang and Antoni~B Chan.
\newblock Learning dynamic memory networks for object tracking.
\newblock In {\em Proceedings of the European conference on computer vision
  (ECCV)}, pages 152--167, 2018.

\bibitem{ostrack}
Botao Ye, Hong Chang, Bingpeng Ma, Shiguang Shan, and Xilin Chen.
\newblock Joint feature learning and relation modeling for tracking: A
  one-stream framework.
\newblock In {\em European Conference on Computer Vision}, pages 341--357.
  Springer, 2022.

\bibitem{SiamAttn}
Yuechen Yu, Yilei Xiong, Weilin Huang, and Matthew~R Scott.
\newblock Deformable siamese attention networks for visual object tracking.
\newblock In {\em Proceedings of the IEEE/CVF conference on computer vision and
  pattern recognition}, pages 6728--6737, 2020.

\bibitem{yuan2022learning}
Di~Yuan, Xiaojun Chang, Zhihui Li, and Zhenyu He.
\newblock Learning adaptive spatial-temporal context-aware correlation filters
  for uav tracking.
\newblock {\em ACM Transactions on Multimedia Computing, Communications, and
  Applications (TOMM)}, 18(3):1--18, 2022.

\bibitem{yuan2023active}
Di~Yuan, Xiaojun Chang, Qiao Liu, Yi~Yang, Dehua Wang, Minglei Shu, Zhenyu He,
  and Guangming Shi.
\newblock Active learning for deep visual tracking.
\newblock {\em IEEE Transactions on Neural Networks and Learning Systems},
  2023.

\bibitem{yuan2022aligned}
Di~Yuan, Xiu Shu, Qiao Liu, and Zhenyu He.
\newblock Aligned spatial-temporal memory network for thermal infrared target
  tracking.
\newblock {\em IEEE Transactions on Circuits and Systems II: Express Briefs},
  70(3):1224--1228, 2022.

\bibitem{yuan2023robust}
Di~Yuan, Xiu Shu, Qiao Liu, Xinming Zhang, and Zhenyu He.
\newblock Robust thermal infrared tracking via an adaptively multi-feature
  fusion model.
\newblock {\em Neural Computing and Applications}, 35(4):3423--3434, 2023.

\bibitem{zhang2017robust}
Le~Zhang and Ponnuthurai~Nagaratnam Suganthan.
\newblock Robust visual tracking via co-trained kernelized correlation filters.
\newblock {\em Pattern Recognition}, 69:82--93, 2017.

\bibitem{UpdateNet}
Lichao Zhang, Abel Gonzalez-Garcia, Joost van~de Weijer, Martin Danelljan, and
  Fahad~Shahbaz Khan.
\newblock Learning the model update for siamese trackers.
\newblock In {\em Proceedings of the IEEE/CVF international conference on
  computer vision}, pages 4010--4019, 2019.

\bibitem{SiamDW}
Zhipeng Zhang and Houwen Peng.
\newblock Deeper and wider siamese networks for real-time visual tracking.
\newblock In {\em Proceedings of the IEEE/CVF Conference on Computer Vision and
  Pattern Recognition}, pages 4591--4600, 2019.

\bibitem{Ocean}
Zhipeng Zhang, Houwen Peng, Jianlong Fu, Bing Li, and Weiming Hu.
\newblock Ocean: Object-aware anchor-free tracking.
\newblock In {\em European Conference on Computer Vision}, pages 771--787.
  Springer, 2020.

\bibitem{1}
Zechu Zhou, Xinyu Zhou, Zhaoyu Chen, Pinxue Guo, Qian-Yu Liu, and Wenqiang
  Zhang.
\newblock Memory network with pixel-level spatio-temporal learning for visual
  object tracking.
\newblock {\em IEEE Transactions on Circuits and Systems for Video Technology},
  2023.

\bibitem{2}
Zikun Zhou, Xin Li, Tianzhu Zhang, Hongpeng Wang, and Zhenyu He.
\newblock Object tracking via spatial-temporal memory network.
\newblock {\em IEEE Transactions on Circuits and Systems for Video Technology},
  32(5):2976--2989, 2021.

\bibitem{9417705}
Hu~Zhu, Hao Peng, Guoxia Xu, Lizhen Deng, Yueying Cheng, and Aiguo Song.
\newblock Bilateral weighted regression ranking model with spatial-temporal
  correlation filter for visual tracking.
\newblock {\em IEEE Transactions on Multimedia}, 24:2098--2111, 2022.

\bibitem{5229254}
Junda Zhu, Yuanwei Lao, and Yuan~F. Zheng.
\newblock Object tracking in structured environments for video surveillance
  applications.
\newblock {\em IEEE Transactions on Circuits and Systems for Video Technology},
  20(2):223--235, 2010.

\bibitem{9914676}
Tianyu Zhu, Markus Hiller, Mahsa Ehsanpour, Rongkai Ma, Tom Drummond, Ian Reid,
  and Hamid Rezatofighi.
\newblock Looking beyond two frames: End-to-end multi-object tracking using
  spatial and temporal transformers.
\newblock {\em IEEE Transactions on Pattern Analysis and Machine Intelligence},
  45(11):12783--12797, 2023.

\bibitem{9318537}
Xue-Feng Zhu, Xiao-Jun Wu, Tianyang Xu, Zhen-Hua Feng, and Josef Kittler.
\newblock Robust visual object tracking via adaptive attribute-aware
  discriminative correlation filters.
\newblock {\em IEEE Transactions on Multimedia}, 24:301--312, 2022.

\bibitem{zhu2017uct}
Zheng Zhu, Guan Huang, Wei Zou, Dalong Du, and Chang Huang.
\newblock Uct: Learning unified convolutional networks for real-time visual
  tracking.
\newblock In {\em Proceedings of the IEEE international conference on computer
  vision workshops}, pages 1973--1982, 2017.

\bibitem{DaSiamRPN}
Zheng Zhu, Qiang Wang, Bo~Li, Wei Wu, Junjie Yan, and Weiming Hu.
\newblock Distractor-aware siamese networks for visual object tracking.
\newblock In {\em Proceedings of the European conference on computer vision
  (ECCV)}, pages 101--117, 2018.

\bibitem{8578162}
Zheng Zhu, Wei Wu, Wei Zou, and Junjie Yan.
\newblock End-to-end flow correlation tracking with spatial-temporal attention.
\newblock In {\em 2018 IEEE/CVF Conference on Computer Vision and Pattern
  Recognition}, pages 548--557, 2018.

\end{thebibliography}
}

\end{document}